\documentclass[a4paper,fleqn]{cas-sc}

\usepackage[authoryear,sort]{natbib}

\usepackage{graphicx}
\usepackage{multirow}
\usepackage{amsmath,amssymb,amsfonts}
\usepackage{mathrsfs}
\usepackage{xcolor}
\usepackage{textcomp}
\usepackage{manyfoot}
\usepackage{booktabs}
\usepackage{algorithm}
\usepackage{algorithmicx}
\usepackage{algpseudocode}
\usepackage{listings}
\usepackage{array}
\usepackage{tabularx}
\usepackage{adjustbox}
\usepackage{float}
\usepackage{placeins}
\hypersetup{
  colorlinks=true,
  citecolor=blue,
  linkcolor=blue,
  urlcolor=blue,
  hypertexnames=false
}

\begin{document}
\let\WriteBookmarks\relax
\def\floatpagepagefraction{1}
\def\textpagefraction{.001}

\shorttitle{}
\shortauthors{Wang et al.}

\title[mode=title]{A Query-Time Framework for Transient 2D Pore-Scale Flow Prediction and Generative Design}


\author[1]{Yiming Wang}

\author[1]{Jiale Zhu}
\cormark[1]
\ead{jiale.zhu@sydney.edu.au}

\author[1]{Zhichen Ye}

\author[2]{Yandong Lv}

\author[3]{Shiqi Wang}

\author[4]{Jinlong Liu}

\author[5]{Yucheng Fan}[orcid=0000-0001-5832-7020]
\cormark[1]
\ead{yucheng.fan@student.unsw.edu.au}

\cortext[1]{Corresponding authors.}


\affiliation[1]{
    organization={School of Civil Engineering, The University of Sydney},
    city={Sydney},
    postcode={2006},
    state={NSW},
    country={Australia}
}

\affiliation[2]{
    organization={MOE Key Laboratory of Soft Soils and Geoenvironmental Engineering, Zhejiang University},
    city={Hangzhou},
    postcode={310058},
    country={China}
}

\affiliation[3]{
    organization={College of Civil Engineering and Architecture, Zhejiang University},
    city={Hangzhou},
    postcode={310058},
    country={China}
}

\affiliation[4]{
    organization={School of Civil Engineering, Southeast University},
    city={Nanjing},
    postcode={211189},
    country={China}
}

\affiliation[5]{
    organization={School of Civil and Environmental Engineering, UNSW Sydney},
    city={Sydney},
    postcode={2052},
    state={NSW},
    country={Australia}
}

\begin{abstract}
Pore-scale flow governs transport and permeability behaviour in porous media engineering applications, yet repeated lattice Boltzmann method (LBM) simulation across many geometries and design queries remains costly for repeated deployment. This study formulates transient pore-scale flow prediction as a geometry-conditioned query-time operator and introduces QSGS-Transient-7606, a benchmark of 7,606 two-dimensional porous structures each paired with 30 logarithmically sampled LBM states. The proposed continuous-time pore-scale flow surrogate model (CT-PoreFlow) integrates topology-aware geometry encoding, compressed spectral mixing, and log-time conditioning with a late-time flux-calibration objective.

On unseen test geometries, CT-PoreFlow achieves a velocity relative $L_2$ of 0.2248 and a terminal permeability error of 12.81\%. Frozen morphology and computed tomography image audits confirm reasonable cross-geometry robustness without fine-tuning. The surrogate is then embedded in an inverse design workflow, screening 9,216 generative adversarial network and diffusion candidates across 18 property targets prior to LBM verification. Guided GAN sampling attains 98.11\% through-connectivity and 72.28\% conditional design success, exceeding diffusion-based generation. The framework unifies transient flow prediction, transport-aware screening, and LBM-verified inverse design for porous media.
\end{abstract}

\begin{highlights}
\item 7,606-geometry benchmark resolves transient pore-scale flow at 30 logarithmic times.
\item Topology-aware CT-PoreFlow improves transient and terminal transport prediction.
\item GAN and diffusion designs are screened by the surrogate and verified with LBM.
\end{highlights}

\begin{keywords}
pore-scale flow \sep lattice Boltzmann method \sep transient surrogate \sep inverse design \sep porous media
\end{keywords}

\maketitle

\section{Introduction}
Pore-scale flow governs the performance of a broad range of engineering systems
in which fluid transport through a disordered solid matrix determines the
macroscopic response of practical interest. In groundwater and contaminant
transport, pore-scale retention and preferential migration influence breakthrough
behaviour \citep{smith2008porescale}; in petroleum recovery, pore geometry and
topology regulate displacement pathways and sweep efficiency
\citep{ju2020porecharacteristics}. Similar pore-scale processes affect the evolution
of permeability and transport pathways during geological CO$_2$ sequestration
\citep{gao2017reactive}, while the microstructure of filtration membranes and porous
flow-battery electrodes controls permeability and fluid or reactant transport
\citep{song2022membrane,li2020electrode}. Across these applications, bulk porosity
alone is insufficient to describe hydraulic behaviour because the connectivity,
constrictions, and organisation of the accessible pore space determine how flow is
distributed through the medium. This distinction is also relevant to geotechnical and mining systems in which bulk void fraction is constrained by solids inventory, density, strength, or
treatment capacity, while pore connectivity and organisation remain adjustable. In heap leaching, compression, moisture, and fines agglomeration can reorganise the connected pore network and create low-permeability or plugged zones \citep{erskine2025heap}. In cemented paste backfill, mixture proportion and curing modify pore size and interconnection, which are closely related to mechanical integrity \citep{huan2021backfill}. In permeable reactive barriers, corrosion products progressively occupy pore space and alter permeability, preferential flow, and contaminant residence time \citep{yang2021prb}. Additively manufactured
rock analogues provide a complementary route in which porosity and internal geometry can be controlled for repeatable structure--transport and structure--failure experiments \citep{zhang2026rockanalog}.

Resolving this structure--transport relationship requires pore-scale flow
simulation, which connects individual pores and throats to permeability, hydraulic
tortuosity, preferential pathways, and transport localisation. The lattice
Boltzmann method (LBM) is well suited to image-based porous media because collision
and streaming are performed on a regular lattice and no-slip solid boundaries can
be imposed directly on a voxelised geometry
\citep{qian1992d2q9,he1997lbm,zou1997boundary}. 
Quartet structure generation set (QSGS) models complement this solver by producing controlled stochastic
microstructures from core-placement and directional-growth parameters
\citep{wang2007qsgs}. Their combination enables systematic investigation of
geometry-dependent transport.

However, the main limitation of LBM in high-throughput pore-scale analysis is the cost of
repeated iterative simulation. Each new geometry requires an independent solve,
while transient analysis further requires the evolving flow field to be resolved
over multiple time states before convergence. Data-driven surrogates can reduce
this cost by learning the geometry-to-flow mapping and providing rapid predictions
for new geometries and query times without repeating the full LBM solution. More
broadly in computational mechanics, high-fidelity simulations have supported
transferable impact-damage prediction and lightweight structural-capacity
surrogates based on Kolmogorov--Arnold networks
\citep{yu2026crossscenario,kong2026lightweight}. Image-to-flow surrogates have reduced this cost for steady fields and scalar transport properties. Early data-driven studies mainly focused on learning steady pore-scale flow
fields or effective transport properties from porous geometry.
\citet{takbiri2020surrogate} developed a fully convolutional surrogate that maps
two-dimensional binary pore images to LBM-resolved velocity fields and then
recovers permeability tensors from the predicted flow fields. 
\citet{santos2020poreflow} later proposed PoreFlow-Net for
three-dimensional digital rocks, where geometric descriptors extracted from
the binary pore space were combined with a convolutional network to improve
steady velocity-field prediction across structurally different porous samples.
\citet{wang2021phyflow} introduced the PhyFlow-HierCAE framework,
embedding discretized Navier--Stokes and continuity information together with
hierarchical regularisation to improve pore-fluid velocity and permeability
prediction. To provide nonlocal transport information,
\citet{zhou2022pore} incorporated a computationally inexpensive coarse velocity field
into a super-resolution network and showed improved fine-scale flow prediction,
particularly for heterogeneous pore structures. \citet{yu2023modified}
further modified U-Net using dense skip connections and attention gates to
capture multilevel flow features in porous media.
In parallel, \citet{wu2018permeability} demonstrated that permeability can be predicted directly
from porous images using convolutional neural networks. These studies established effective routes for
steady field reconstruction and property prediction, but most formulations
remain centred on converged flow states or scalar transport quantities rather
than direct query-time reconstruction of the full transient pore-scale
evolution. Addressing this gap means modelling the pre-steady evolution itself, which raises issues
distinct from steady-state mapping.

Transient pore-scale flow is more difficult to approximate than a converged
steady state because pressure redistribution and velocity development occur
simultaneously but not necessarily at the same rate. The evolution also depends
strongly on pore connectivity. Geometries with similar porosity or local aperture
may exhibit different relaxation behaviour when narrow throats, dead-end pores,
or long inlet-to-outlet pathways are present. Recurrent models such as ConvLSTM
can describe temporal evolution by propagating information from one state to the
next \citep{shi2015convlstm,kumar2025adcrnn}, but this sequential formulation is
naturally tied to the temporal discretisation used during training. Other studies
have instead sought direct surrogates for repeated transient or multi-query flow
analysis. \citet{pinheiro2026surrogate} compared reduced-order and fully convolutional models
for evolving rock--fluid interaction and introduced grid-size-invariant surrogates
that were trained on subdomains and evaluated on larger unseen domains.
At the reservoir scale, \citet{wen2022ufno} combined Fourier
layers with a U-Net branch in U-FNO to predict space--time pressure and saturation
fields for three-dimensional multiphase $\mathrm{CO}_2$--water flow. These studies demonstrate that time-dependent porous-flow
responses can be learned without relying exclusively on conventional sequential
simulation.

The challenge becomes more demanding at the pore scale, where transient prediction
must preserve both temporal evolution and the detailed influence of pore boundaries
and connected transport pathways. \citet{mesgarpour2023transient} combined high-fidelity flow
simulation with physics-informed learning to predict transient three-dimensional
flow through fibrous porous media, while \citet{kim2026prtdeeponet}
introduced geometry and geodesic information into DeepONet to interpolate
time-dependent pore-scale reactive transport. However, this formulation interpolates a concentration field
under a velocity field that must already be supplied, rather than predicting the
velocity and pressure response of a new geometry directly from rest. This leaves an open need for a non-sequential geometry-to-field formulation capable of representing transient pore-scale evolution efficiently across repeated time queries.

Neural operators provide a suitable framework
for this purpose because they learn mappings between input conditions and physical
fields \citep{kovachki2023neuraloperator,lu2021deeponet}. Fourier-based operators
further provide a mechanism for exchanging information over long spatial distances,
which is useful when transport depends on connected pathways extending across the
sample \citep{li2021fno,song2025multigrid}. For pore-scale flow, however, global
information alone is insufficient. Narrow throats and no-slip boundaries require
accurate local reconstruction, whereas permeability and hydraulic connectivity
depend on the organisation of transport pathways across the entire domain. Therefore, an
effective transient surrogate needs to combine local pore-scale resolution
with global transport information while remaining efficient for repeated engineering
queries.

However, accurate forward prediction addresses only one side of the engineering
problem. In many porous-media applications, the practical objective is not merely
to evaluate a given geometry, but to identify pore structures that achieve prescribed
transport properties. Recent studies have therefore begun to combine learned
structure--property relations with generative or optimisation-based inverse design.
For example, \citet{nguyen2026inverse} coupled a property-conditioned generative model with a
surrogate predictor to design porous microstructures targeting porosity and
permeability, while related approaches have used
variational, flow-based, and diffusion models to explore property-guided
microstructure generation. Generative learning has also supported engineering
data augmentation; for example, CTGAN-generated samples have been coupled with
stacked machine-learning models for recycled-concrete strength prediction
\citep{liu2026synthetic}. A remaining difficulty is that a generated geometry
that matches a surrogate-predicted target may still be hydraulically invalid,
disconnected, or inaccurate when returned to the original numerical solver.
Consequently, there remains a need for an inverse-design workflow that links
pore-scale field prediction to engineering transport quantities, uses the surrogate
for efficient candidate screening, and retains the high-fidelity solver for final
LBM verification.

\begin{figure}[pos=htbp]
\centering
\includegraphics[width=\textwidth,height=0.70\textheight,keepaspectratio]{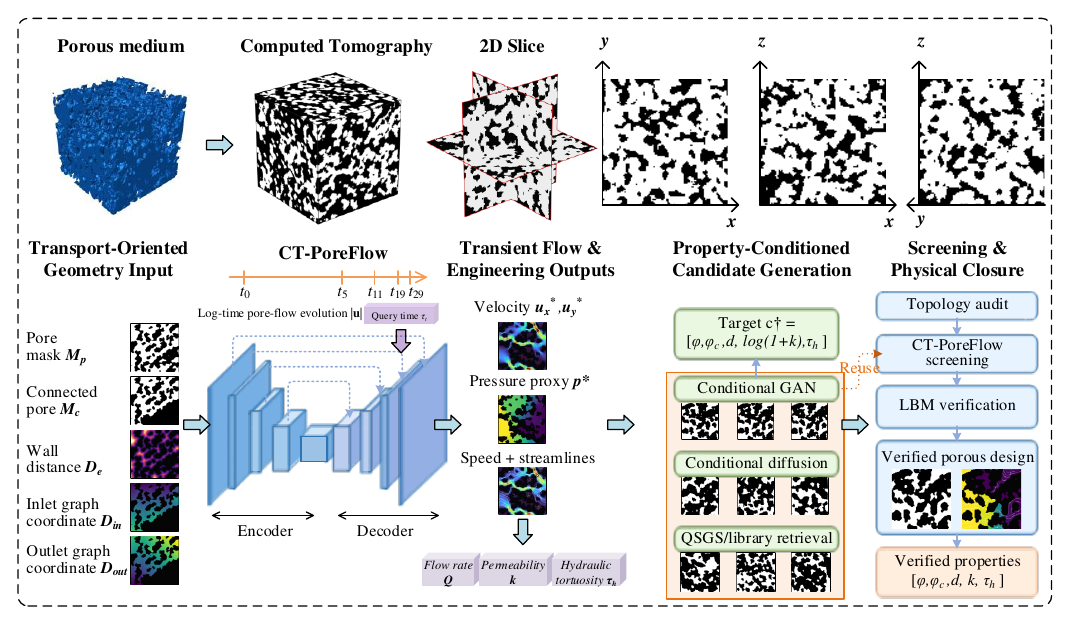}
\caption{The computational framework.}
\label{fig:task_pipeline}
\end{figure}

Hence, to address these research gaps, this study develops an integrated computational framework
that connects transient pore-scale flow prediction with transport-oriented porous
structure design. As shown in Fig.~\ref{fig:task_pipeline}, a large-scale QSGS--LBM benchmark is first constructed to resolve
the evolution of velocity and pressure fields from rest toward steady flow across
thousands of statistically distinct porous geometries. Based on these data,
continuous-time pore-scale flow surrogate model (CT-PoreFlow) is developed as a geometry-conditioned query-time surrogate that
combines local pore information with global transport-path representation, allowing
the flow state to be predicted directly at a requested time without sequentially
advancing through the preceding states. The predicted fields are further used to
recover engineering quantities including flow rate, permeability, and hydraulic
tortuosity, so that model performance is evaluated in terms of both field fidelity
and transport accuracy. Finally, the forward surrogate is embedded into an inverse
design workflow in which candidate porous structures are generated by cGAN or diffusion model, screened according to their predicted transport response, and returned to the LBM solver for final LBM verification. Together, these developments establish a unified computational framework that
connects transient pore-scale flow prediction with engineering transport analysis
and inverse porous-structure design. This framework provides a foundation for computationally efficient and physics-verified inverse design of porous media with prescribed transport properties.

\FloatBarrier

\section{Methodology}
\label{sec:method}
\subsection{Pore-scale flow simulation}
\label{sec:flow_problem}

The pore-scale simulations are designed to resolve how fluid develops through
a heterogeneous pore network and how the resulting local flow field gives rise
to effective transport properties. A pressure difference is imposed across each
porous structure, driving the fluid from rest toward a converged seepage state.
During this process, the evolving velocity and pressure fields provide the basis
for evaluating flow rate, permeability, and hydraulic tortuosity. The two-dimensional computational domain is a $100\times100$ lattice, $\Omega=\{1,\ldots,H\}\times\{1,\ldots,W\}$ with $H=W=100$, partitioned into pore and solid regions,
\begin{equation}
\Omega=\Omega_p\cup\Omega_s,
\qquad
\Omega_p\cap\Omega_s=\varnothing.
\label{eq:pore_domain}
\end{equation}
The solid mask $M_s$ equals one on $\Omega_s$ and zero on $\Omega_p$; the pore mask is $M_p=1-M_s$. These masks first define the numerical domain and boundary locations. Their later use as geometry channels does not change this physical meaning.

At the continuum scale, the weakly compressible macroscopic response recovered by the LBM is described by mass conservation and momentum balance,
\begin{equation}
\frac{\partial\rho}{\partial t}
+\nabla\cdot(\rho\mathbf{u})=0,
\label{eq:mass_conservation}
\end{equation}
\begin{equation}
\rho\left(
\frac{\partial\mathbf{u}}{\partial t}
+\mathbf{u}\cdot\nabla\mathbf{u}
\right)
=-\nabla p+\mu\nabla^2\mathbf{u},
\label{eq:navier_stokes}
\end{equation}
where $\mathbf{u}=(u_x,u_y)$ is velocity, $p$ is pressure, $\rho$ is density, and $\mu$ is dynamic viscosity. The engineering problem is low-speed pressure-driven seepage imposed in the LBM through a small inlet--outlet lattice-density difference. Under the low-Mach operating regime enforced in this study, density fluctuations remain small and the recovered flow approaches the incompressible limit,
\begin{equation}
\nabla\cdot\mathbf{u}\simeq0.
\label{eq:low_mach_limit}
\end{equation}
In this study, the flow is solved using the D2Q9-BGK lattice Boltzmann formulation,
where D2Q9 represents a two-dimensional lattice with nine discrete particle
velocities, and the BGK collision operator relaxes the particle distributions toward
local equilibrium through a single relaxation time
\citep{qian1992d2q9,he1997lbm}. Under the low-Mach regime considered here, this
formulation recovers the weakly compressible Navier--Stokes behaviour described by
Eqs.~\ref{eq:mass_conservation}--\ref{eq:low_mach_limit}.

The local LBM fields are converted into engineering observables using fixed lattice definitions. Total and through-connected porosities are
\begin{equation}
\phi=\frac{|\Omega_p|}{|\Omega|},
\label{eq:porosity}
\end{equation}
\begin{equation}
\phi_c=\frac{|\Omega_c|}{|\Omega|},
\label{eq:connected_porosity}
\end{equation}
where $\Omega_c\subseteq\Omega_p$ is the four-neighbour pore region reachable from both the inlet and outlet. The inverse-design property vector also uses the code-defined dead-pore ratio
\begin{equation}
d=1-\frac{\phi_c}{\phi}.
\label{eq:dead_pore_ratio}
\end{equation}

Flow is aligned with lattice axis 0. Let $\Gamma_{\mathrm{in}}$ and $\Gamma_{\mathrm{out}}$ denote the two open measurement rows. The LBM reference solver uses the fluid buffer rows immediately adjoining the core, whereas predicted-field evaluation uses the first and last stored sample rows. Both implementations sum streamwise velocity over their respective open nodes,
\begin{equation}
Q_{\mathrm{in}}=\sum_{j\in\Gamma_{\mathrm{in}}}u_x(j),
\qquad
Q_{\mathrm{out}}=\sum_{j\in\Gamma_{\mathrm{out}}}u_x(j),
\qquad
Q=\frac{Q_{\mathrm{in}}+Q_{\mathrm{out}}}{2},
\label{eq:boundary_flow_rate}
\end{equation}
and defines the two-dimensional Darcy velocity by
\begin{equation}
u_D=\frac{Q}{W}.
\label{eq:darcy_velocity}
\end{equation}
With a positive streamwise pressure drop $\Delta p=p_{\mathrm{in}}-p_{\mathrm{out}}>0$, the Darcy relation is
\begin{equation}
u_D=\frac{k}{\mu_{\mathrm{ref}}}\frac{\Delta p}{H},
\label{eq:darcy_law}
\end{equation}
and therefore
\begin{equation}
k=\frac{\mu_{\mathrm{ref}}H u_D}{\Delta p}.
\label{eq:darcy_permeability}
\end{equation}
The LBM reference solver evaluates this relation as
\begin{equation}
k_{\mathrm{LBM}}=\frac{\mu_{\mathrm{ref}}H u_D}{\Delta p_{\mathrm{LBM}}},
\qquad
\mu_{\mathrm{ref}}=\rho_{\mathrm{ref}}\nu,
\qquad
\rho_{\mathrm{ref}}=\frac{\overline{\rho}_{\Gamma_{\mathrm{in}}}+\overline{\rho}_{\Gamma_{\mathrm{out}}}}{2}.
\label{eq:permeability}
\end{equation}
Here $\Delta p_{\mathrm{LBM}}$ is measured on the two solver-side sample faces. For prediction--reference comparison, the same Darcy expression is applied to both tensors using the imposed lattice pressure difference $\Delta p_0$, as defined in Eq.~\ref{eq:pressure_drop} below, and the implemented approximation $\rho_{\mathrm{ref}}\simeq1$. The two code paths are therefore reproducible without interrupting the engineering definition.

Hydraulic tortuosity is calculated from the resolved terminal velocity field as
\begin{equation}
\tau_h=
\frac{\left\langle|\mathbf{u}|\right\rangle_{\Omega_p}}
{\left|\left\langle u_x\right\rangle_{\Omega_p}\right|}
=
\frac{\sum_{\mathbf{x}\in\Omega_p}|\mathbf{u}(\mathbf{x})|}
{\left|\sum_{\mathbf{x}\in\Omega_p}u_x(\mathbf{x})\right|}.
\label{eq:hydraulic_tortuosity}
\end{equation}

The QSGS constructs statistically distinct binary microstructures while retaining direct control over phase fraction and directional growth \citep{wang2007qsgs}. Target porosity is sampled in eight width-0.05 intervals spanning 0.30--0.70. The solid-core probability and four axial growth probabilities are 0.01. The four diagonal probabilities are 0.03 except in the $(+x,-y)$ direction, where the value is 0.05. A zero-padded $3\times3$ binary median filter is applied twice after growth.

Each task is retried until the filtered porosity lies within its assigned interval and a pore component connects inlet to outlet. Connectivity is determined with four-neighbour labelling before any LBM solve. Isolated cavities remain part of the binary geometry, and cul-de-sac branches attached to a spanning component remain within that component. This preserves void storage and local pressure effects instead of replacing the generated structure with an idealized transport backbone. A candidate is accepted into the dataset only after the connectivity audit
and the numerical quality-control criteria described below.

\begin{figure}[pos=htbp]
\centering
\includegraphics[width=\textwidth,height=0.72\textheight,keepaspectratio]{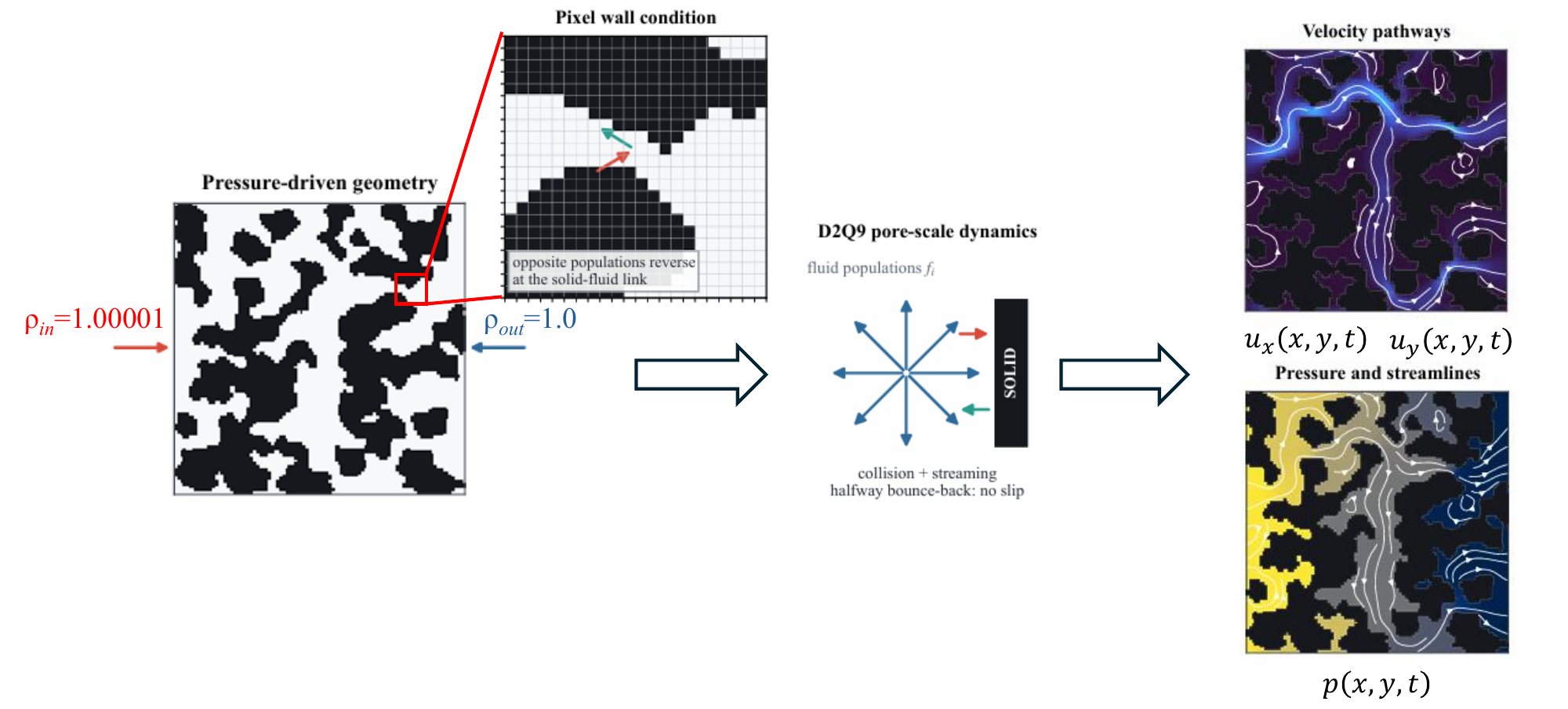}
\caption{From pixel-resolved geometry to Darcy-scale transport.}
\label{fig:lbm_principle}
\end{figure}

As shown in Fig.~\ref{fig:lbm_principle}, every geometry uses the same D2Q9-BGK configuration: $\rho_{\mathrm{in}}=1.00001$, $\rho_{\mathrm{out}}=1.0$, relaxation time $\tau_{\mathrm{LBM}}=1.0$, lattice kinematic viscosity $\nu=(\tau_{\mathrm{LBM}}-0.5)/3$, five fluid buffer rows at each pressure boundary, and bounce-back on solid nodes. The pressure-driven seepage is imposed through density boundary conditions following standard lattice-BGK formulations \citep{qian1992d2q9,zou1997boundary}. 

\FloatBarrier

For population $f_i$ with lattice velocity $\mathbf{c}_i$, collision and streaming are
\begin{equation}
f_i(\mathbf{x}+\mathbf{c}_i,t+1)
=f_i(\mathbf{x},t)
-\frac{1}{\tau_{\mathrm{LBM}}}
\left[f_i(\mathbf{x},t)-f_i^{\mathrm{eq}}(\mathbf{x},t)\right].
\label{eq:lbgk}
\end{equation}
The equilibrium distribution is
\begin{equation}
f_i^{\mathrm{eq}}
=w_i\rho\left[
1+3\mathbf{c}_i\cdot\mathbf{u}
+\frac{9}{2}(\mathbf{c}_i\cdot\mathbf{u})^2
-\frac{3}{2}|\mathbf{u}|^2
\right],
\label{eq:equilibrium_distribution}
\end{equation}
where $w_i=4/9$ for the rest population, $1/9$ for the four axial directions, and $1/36$ for the four diagonals. Density and momentum are recovered from population moments,
\begin{equation}
\rho=\sum_i f_i,
\qquad
\rho\mathbf{u}=\sum_i f_i\mathbf{c}_i.
\label{eq:macroscopic_moments}
\end{equation}
For D2Q9, the lattice pressure relation is
\begin{equation}
p=c_s^2\rho,
\qquad
c_s^2=\frac{1}{3},
\label{eq:pressure_relation}
\end{equation}
and the imposed pressure difference is therefore
\begin{equation}
\Delta p_0=c_s^2(\rho_{\mathrm{in}}-\rho_{\mathrm{out}})
=\frac{\rho_{\mathrm{in}}-\rho_{\mathrm{out}}}{3}.
\label{eq:pressure_drop}
\end{equation}
For stored reference permeability, the high-fidelity solver instead measures $\Delta p_{\mathrm{LBM}}$ from the mean densities on the fluid rows adjacent to the sample, consistently with Eq.~\ref{eq:permeability}.
The learning and visualisation target is the normalised density-based pressure proxy
\begin{equation}
p^*=\frac{\rho-\rho_{\mathrm{out}}}
{\rho_{\mathrm{in}}-\rho_{\mathrm{out}}}.
\label{eq:density_pressure_proxy}
\end{equation}

The simulation starts from rest and is checked every 500 iterations. If $n$ denotes the current check and $n-1$ the preceding check, the implemented velocity residual is
\begin{equation}
r_u^{(n)}=
\frac{
\left[\sum_{\mathbf{x}\in\Omega_f}
\left|\mathbf{u}^{(n)}(\mathbf{x})-\mathbf{u}^{(n-1)}(\mathbf{x})\right|^2\right]^{1/2}}
{
\left[\sum_{\mathbf{x}\in\Omega_f}|\mathbf{u}^{(n)}(\mathbf{x})|^2\right]^{1/2}+10^{-30}},
\label{eq:velocity_residual}
\end{equation}
where $\Omega_f$ contains fluid nodes in the padded solver lattice. The reference-solver flux imbalance and permeability change are
\begin{equation}
e_m^{\mathrm{QC}}=
\frac{|Q_{\mathrm{in}}-Q_{\mathrm{out}}|}
{\max(|Q_{\mathrm{in}}|,|Q_{\mathrm{out}}|,\epsilon)},
\label{eq:mass_imbalance_qc}
\end{equation}
\begin{equation}
r_k^{(n)}=
\frac{|k^{(n)}-k^{(n-1)}|}
{\max(|k^{(n)}|,|k^{(n-1)}|,\epsilon)}.
\label{eq:permeability_residual}
\end{equation}
Here $\epsilon$ is the floating-point machine epsilon used by the implementation.
Weak-compressibility control is expressed by
\begin{equation}
\mathrm{Ma}_{\max}=\frac{\max_{\Omega_f}|\mathbf{u}|}{c_s}.
\label{eq:mach_number}
\end{equation}
A realisation is retained only when $r_u<10^{-4}$, $r_k<10^{-4}$, $e_m^{\mathrm{QC}}<10^{-3}$, all states are finite, and $\mathrm{Ma}_{\max}<0.10$. The solve is capped at 50,000 iterations. Thirty fields are requested on the logarithmic schedule specified in Section~\ref{sec:dataset_geometry}. When convergence occurs before a requested late time, the converged field is copied to the remaining storage positions and marked by a steady-fill indicator.

\subsection{Dataset and Geometry Representation}
\label{sec:dataset_geometry}

QSGS-Transient-7606 contains 7,606 accepted geometries spanning porosity 0.30--0.70. Thirty states were stored on a logarithmically distributed schedule between iteration 0 and 50,000, giving 228,180 geometry--time records. The geometry-disjoint split contains 6,078 training, 764 validation, and 764 test structures. All static channels and all 30 states of a geometry remain in the same split, and normalisation statistics are calculated from training data only.

\begin{figure}[pos=htbp]
\centering
\includegraphics[width=\textwidth,height=0.74\textheight,keepaspectratio]{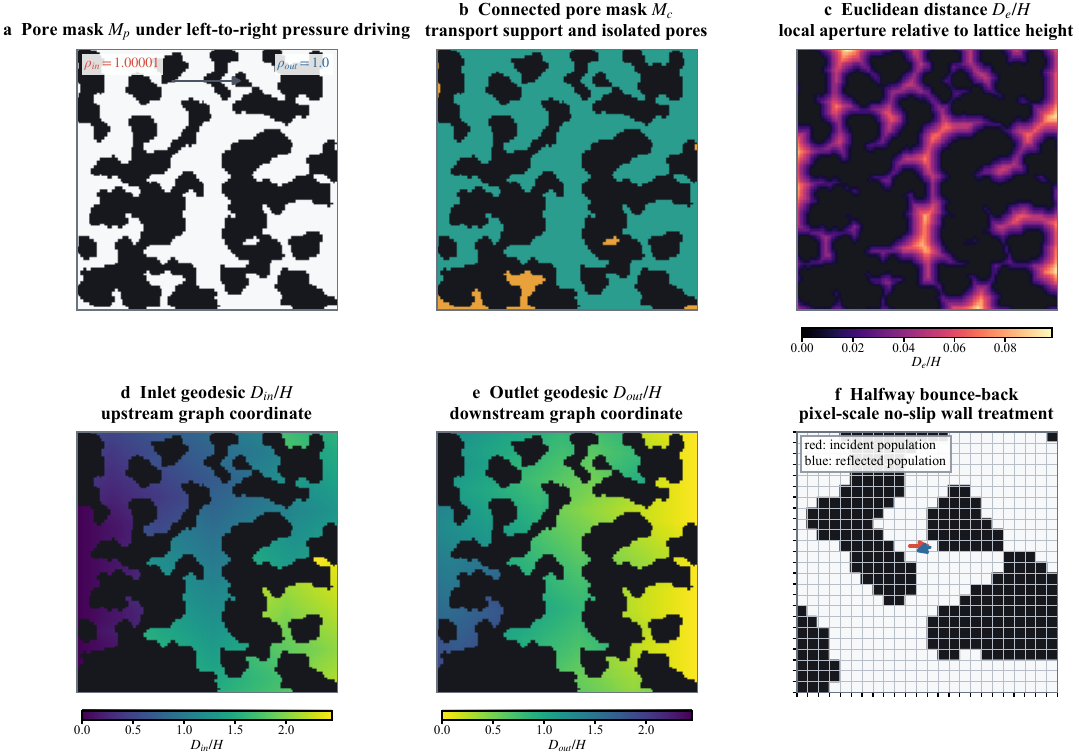}
\caption{Pixel-resolved geometry and transport-oriented topology representation. (a) Pore mask $M_p$ under left-to-right pressure driving. (b) Connected-pore mask $M_c$: teal is through-connected pore space, orange is nonspanning pore space, and black is solid. (c) Euclidean wall distance $D_e/H$. (d,e) Inlet- and outlet-referenced four-neighbour geodesic distances. Panels (a--e) form $\mathbf{X}_G$. (f) Halfway bounce-back at the same pixel-scale interface.}
\label{fig:pixel_geometry}
\label{fig:topology}
\end{figure}

The native pixel lattice is retained without contour smoothing. CT-PoreFlow represents each geometry by
\begin{equation}
\mathbf{X}_G=[M_p,M_c,D_e,D_{\mathrm{in}},D_{\mathrm{out}}]
\in\mathbb{R}^{5\times H\times W}.
\label{eq:input}
\end{equation}
As shown in Fig.~\ref{fig:pixel_geometry}, $M_p$ defines the fluid domain; $M_c$ identifies the inlet--outlet through-connected region; the Euclidean distance transform $D_e/H$ describes local aperture relative to the nearest solid boundary \citep{maurer2003edt}; and $D_{\mathrm{in}}$ and $D_{\mathrm{out}}$ provide inlet- and outlet-referenced shortest-path positions on the four-neighbour pore graph. The graph distances are divided by $\max(H,W)=100$ and set to zero outside $M_c$. Values may exceed one when the connected path is longer than the sample width. Thus, the five channels encode local aperture, hydraulic connectivity, and global transport position.

Multi-source breadth-first traversals start from all open inlet and outlet nodes. Their reachability intersection defines $M_c$, while directed shortest-path lengths give $D_{\mathrm{in}}$ and $D_{\mathrm{out}}$. Attached cul-de-sac branches remain connected when reachable from both boundary traversals. All channels are deterministic functions of geometry and boundary orientation.

\FloatBarrier

A single velocity scale $s_u=5.024\times10^{-6}$ lattice units is estimated from the training split as the 99.5th percentile of a deterministic pore-speed subsample. Both components use the same scale,
\begin{equation}
u_x^*=\frac{u_x}{s_u},
\qquad
u_y^*=\frac{u_y}{s_u},
\label{eq:velocity_normalisation}
\end{equation}
while the normalised pressure proxy $p^*$ follows Eq.~\ref{eq:density_pressure_proxy}. Values above the velocity percentile are not clipped. A supervised item contains one $5\times100\times100$ static geometry tensor, one scalar query time, and one $3\times100\times100$ target $[u_x^*,u_y^*,p^*]$. Stored target values in solid regions are zero.

\subsection{Transient Pore-Flow Surrogate model}
\label{sec:transient_surrogate}

\begin{figure}[pos=htbp]
\centering
\includegraphics[width=\textwidth,height=0.70\textheight,keepaspectratio]{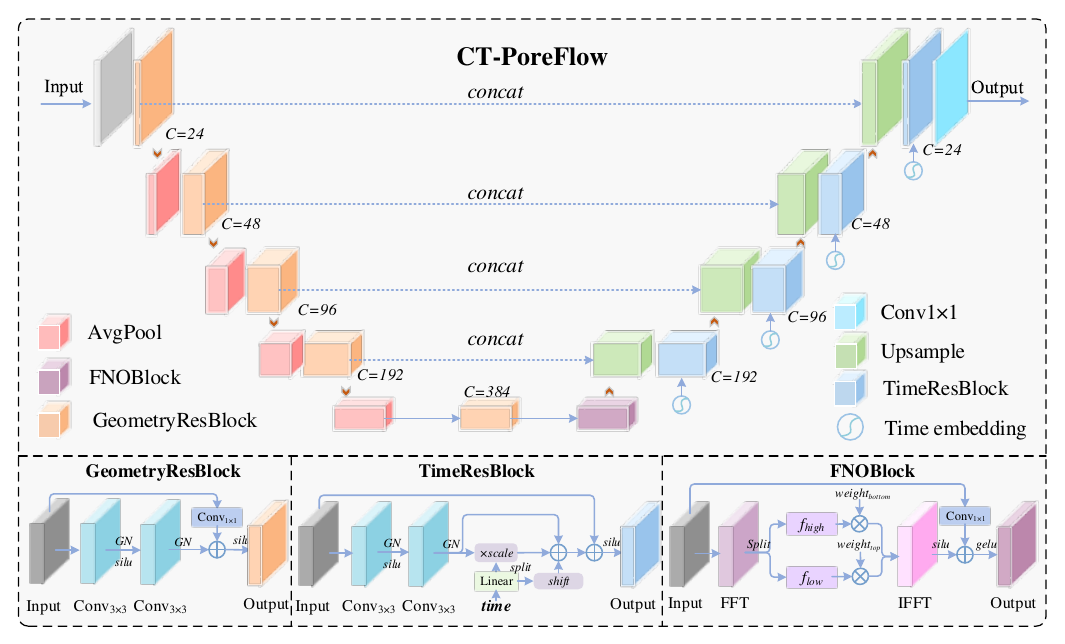}
\caption{CT-PoreFlow computational graph.}
\label{fig:architecture}
\end{figure}

To capture the transient pore-scale response directly from geometry and query time,
we develop the CT-PoreFlow surrogate.
As illustrated in Fig.~\ref{fig:architecture}, CT-PoreFlow follows a local--global
encoder--decoder architecture in which transport-oriented geometry features are
first encoded at multiple spatial scales, global pore-network information is mixed
at the compressed representation, and time-conditioned decoding reconstructs the
requested velocity and pressure fields. Only after the reference problem and data representation have been defined is the geometry-to-field operator introduced. For geometry $G$ and requested LBM iteration $t$,
\begin{equation}
\widehat{\mathbf{Y}}(G,t)=
\mathcal{F}_{\theta}\!\left(\mathbf{X}_G,\tau_t\right),
\qquad
\mathbf{Y}=[u_x^*,u_y^*,p^*],
\label{eq:surrogate_operator}
\end{equation}
with logarithmic query coordinate
\begin{equation}
\tau_t=\frac{\log(1+t)}{\log(1+t_{\max})},
\qquad t_{\max}=50{,}000.
\label{eq:query_time}
\end{equation}
The surrogate is evaluated directly at a requested time; it does not recursively advance the preceding prediction. The logarithmic coordinate allocates more resolution to the densely sampled early transient while retaining the terminal state at $\tau_t=1$.

\subsubsection{GeometryResBlock Encoder}

The GeometryResBlock encoder contains residual stages of widths $C$, $2C$, $4C$, and $8C$, followed by a $16C$ bottleneck, with $C=24$. Each GeometryResBlock uses two $3\times3$ convolutions, Group Normalisation \citep{wu2018groupnorm}, SiLU activations, and a $1\times1$ residual projection when the channel width changes. The AvgPool operation maps $100\times100$ to $50\times50$, $25\times25$, $12\times12$, and $6\times6$. Same-scale tensors are retained for decoder skips following the local-detail principle of U-Net \citep{ronneberger2015unet}. Because the encoder depends only on $\mathbf{X}_G$, its representation can be reused for multiple time queries.

\subsubsection{FNOBlock and Log-Time Conditioning}

At the $6\times6$ bottleneck, the FNOBlock implements compressed spectral mixing across the full sample. A real two-dimensional FFT is multiplied by learned complex weights for three retained modes in each direction, then returned to physical space. A parallel $1\times1$ convolution provides local channel mixing. Their normalised sum is activated and added residually,
\begin{equation}
z'=z+\sigma\!\left[
\operatorname{GN}\!\left(
\mathcal{F}^{-1}(W_k\mathcal{F}(z))+W_0z
\right)\right].
\label{eq:spectral}
\end{equation}
The FNOBlock \citep{li2021fno} acts only on the compressed representation, while wall and throat detail remains available through convolutional skips. As an implementation note, the FFT is evaluated in FP32 within otherwise mixed-precision training.

The scalar $\tau_t$ is embedded into 96 dimensions using sine and cosine functions at exponentially spaced frequencies from 1 to $10^{-4}$, followed by two linear layers and SiLU activation \citep{tancik2020fourier}. Each TimeResBlock maps the embedding to feature-wise scale and shift parameters \citep{perez2018film},
\begin{equation}
h'=h\odot[1+\gamma(\tau_t)]+\beta(\tau_t).
\label{eq:film}
\end{equation}
This introduces time dependence without appending a spatially constant time image at every resolution.

\subsubsection{Pore field reconstruction}

Four bilinear Upsample stages match the encoder resolutions, concatenate the corresponding skip tensors, and reconstruct local velocity and density-proxy fields with TimeResBlocks. Time modulation from Eq.~\ref{eq:film} is applied at every decoder scale. A final Conv1$\times$1 output head produces three channels, followed by the hard pore mask,
\begin{equation}
\widehat{\mathbf{Y}}(G,t)=
M_p\odot D\!\left(E(\mathbf{X}_G),\tau_t\right).
\label{eq:query_decoder}
\end{equation}

\FloatBarrier

\label{sec:transport_objectives}

The reconstructed fields are trained to preserve local numerical detail and remain useful for Darcy-scale transport evaluation. For sample $i$, lattice node $x$, target channel $c$, and sample weight $w_i$, the pore-masked Smooth L1 data term is
\begin{equation}
\mathcal{L}_{\mathrm{data}}=
\frac{\sum_{i,x,c}w_iM_p(x)\,
\operatorname{SmoothL1}(\widehat{Y}_{i,c}(x)-Y_{i,c}(x))}
{\sum_{i,x,c}w_iM_p(x)}.
\label{eq:data_loss}
\end{equation}
Smooth L1 retains quadratic behaviour near zero while limiting the effect of large residuals \citep{huber1964robust}.

The gradient term compares first differences only when both adjacent nodes are fluid,
\begin{equation}
\mathcal{L}_{\mathrm{grad}}=
\frac{1}{2}\left(
\|\Delta_x\widehat{Y}-\Delta_xY\|_{1,\Omega_p}
+\|\Delta_y\widehat{Y}-\Delta_yY\|_{1,\Omega_p}
\right).
\label{eq:gradient_loss}
\end{equation}
Edge masking prevents the intentional discontinuity at the hard pore mask from being treated as a prediction error and preserves gradients around narrow pore throats.

For boundary $b\in\{\mathrm{in},\mathrm{out}\}$, let $q_b(Y)$ be the sum of normalised $u_x^*$ over boundary pore nodes. With
$s_i=\max(|[q_{\mathrm{in}}(Y_i)+q_{\mathrm{out}}(Y_i)]/2|,0.05)$, the flux-calibration term is
\begin{equation}
\mathcal{L}_{\mathrm{flux}}=
\frac{
\sum_i\tau_i^4\,\frac{1}{2}\sum_b
\operatorname{SmoothL1}_{\beta=0.1}\!\left(
\frac{q_b(\widehat{Y}_i)-q_b(Y_i)}{s_i}
\right)}
{\sum_i\tau_i^4}.
\label{eq:flux_loss}
\end{equation}
The time weight emphasizes developed flow, and the scale floor prevents unstable relative penalties near rest. The total objective is
\begin{equation}
\mathcal{L}=
\mathcal{L}_{\mathrm{data}}
+0.05\mathcal{L}_{\mathrm{grad}}
+0.002\mathcal{L}_{\mathrm{flux}}.
\label{eq:total_loss}
\end{equation}
Thus, $\mathcal{L}_{\mathrm{data}}$ controls normalised field fidelity, $\mathcal{L}_{\mathrm{grad}}$ preserves local pore/throat gradients, and $\mathcal{L}_{\mathrm{flux}}$ calibrates a late-time Darcy-relevant boundary quantity.

\subsection{Training and validation}
\label{sec:training_evaluation}

Training and checkpoint selection were kept common wherever the model formulation permitted. AdamW \citep{loshchilov2019adamw}, cosine learning-rate decay \citep{loshchilov2017sgdr}, automatic mixed precision, and gradient-norm clipping at 1.0 were used. Validation uses 256 validation geometries, randomly selected without replacement
from the validation split, evaluated at indices $\{0,5,10,15,20,25,29\}$.

All baseline models received the same five-channel geometry tensor. Time-U-Net is a
convolutional encoder--decoder with sinusoidal time modulation. Time-FNO concatenates these channels with normalised $x$ and $y$ coordinates and a time plane. It uses width 48, 16 modes per direction, and four spectral blocks. DeepONet combines a convolutional geometry branch with an $(x,y,\tau_t)$ trunk and 64 basis functions per output channel \citep{lu2021deeponet}.The remaining controlled configurations are named by the altered component: Without
directed geodesic channels, Without FNOBlock, Linear time coordinate, Field loss
only, and Terminal-anchored steady-residual decoder. CT-PoreFlow's own decoder (Eq.~\ref{eq:query_decoder}) evaluated before flux
calibration is applied, denoted the Direct transient decoder, is also reported to
separate the effect of $\mathcal{L}_{\mathrm{flux}}$ from the decoder formulation itself.

Evaluation considered field fidelity together with engineering transport recovery. Pore-masked MAE and RMSE are reported for $u_x^*$, $u_y^*$, and $p^*$, along with two-component velocity relative $L_2$ and physical speed MAE after inverse scaling. Velocity relative $L_2$ is defined only when the pore-restricted target norm exceeds $10^{-6}$; the zero initial field contributes to absolute error and is excluded from this relative metric. These measures are accompanied by $Q$, $u_D$, $k_{\mathrm{eval}}$, and $\tau_h$ as defined in Section~\ref{sec:flow_problem}. Because $H$, $\nu$, and $\Delta p$ are fixed, $k_{\mathrm{eval}}$ is a constant multiple of $Q$ and their relative errors are not treated as independent gains.

For predicted-field reporting, mass imbalance uses the symmetric diagnostic
\begin{equation}
e_m^{\mathrm{eval}}=
\begin{cases}
\dfrac{|Q_{\mathrm{in}}-Q_{\mathrm{out}}|}
{\tfrac{1}{2}(|Q_{\mathrm{in}}|+|Q_{\mathrm{out}}|)},
& \tfrac{1}{2}(|Q_{\mathrm{in}}|+|Q_{\mathrm{out}}|)>\epsilon_{\mathrm{eval}},\\[6pt]
\mathrm{NaN}, & \text{otherwise},
\end{cases}
\qquad \epsilon_{\mathrm{eval}}=10^{-12}.
\label{eq:mass_imbalance_evaluation}
\end{equation}
This is a validity test rather than a training regularizer, and is distinct from the
max-denominator acceptance gate $e_m^{\mathrm{QC}}$ in Eq.~\ref{eq:mass_imbalance_qc}.
At exact rest the reference flux scale is zero, so the value is stored as NaN and
excluded from finite-value aggregates; reported comparisons use the terminal state.
Divergence is the mean absolute centred-difference divergence on interior nodes whose
four axial neighbours are fluid. Exact masking guarantees zero solid-node velocity by
construction. Same-time relative flux, terminal permeability, and hydraulic-tortuosity
errors are likewise retained only when their reference denominators exceed
$10^{-12}$; reported flux and permeability comparisons use the terminal state.

Dynamic metrics are aggregated within each geometry, and terminal metrics contribute one paired value per geometry, giving $n=764$. Confidence intervals use 10,000 paired geometry-level bootstrap resamples \citep{efron1979bootstrap}; comparisons use two-sided Wilcoxon signed-rank tests, matched rank-biserial effects, and Holm--Bonferroni correction at $\alpha=0.05$ \citep{holm1979multiple}. Errors are stratified by total porosity, connected porosity, dead-pore ratio, hydraulic tortuosity, and LBM convergence iteration.

Two additional audits examined temporal interpolation and morphology transfer. The principal experiment trains on all 30 stored times. A separate interpolation audit trains dedicated CT-PoreFlow and Time-U-Net models on 20 indices and evaluates the absent indices $\{2,4,7,9,12,14,17,19,22,27\}$ on all test geometries. Off-grid visualisation queries have no LBM labels and are not used as accuracy evidence. Model-only latency is measured after warm-up with synchronized CUDA events; end-to-end latency additionally includes loading and preprocessing. Numerical tensors are exported before rendering, and all displayed speeds use training-derived scales rather than per-model RGB normalisation.

Morphology out-of-distribution (OOD) testing was conducted using two
complementary datasets: rule-based synthetic geometries and pore
structures extracted from real computed tomography images. All
models were frozen during OOD evaluation, and no fine-tuning,
recalibration, or OOD-specific parameter optimisation was performed.

Following \citet{he2024oodporous}, the rule-based OOD dataset contains 24 geometries comprising circular arrays,
snowflake obstacles, tree-like obstacles, and staggered baffles.
Each family contains two porosity levels and three seeded replicates.
These geometries were generated independently of the QSGS procedure
used to construct the training dataset.

The computed tomography-derived OOD dataset contains eight supplied binary two-dimensional
slices from one segmented computed tomography volume. In the source images, white denotes pore
and black denotes solid. The $200\times200$ masks were mapped to the fixed
$100\times100$ model lattice by non-overlapping $2\times2$ area averaging with
a pore threshold of 0.5; no topology repair was applied. Array orientation was
then mapped so that solver axis 0 and every displayed field follow the prescribed
left-to-right pressure gradient. Each mask underwent the same connected-pore,
Euclidean-distance, and directed-geodesic preprocessing as the training data,
followed by a terminal D2Q9-BGK LBM reference solve. The terminal metric is
\begin{equation}
e_T=
\frac{\|[\widehat{\mathbf{u}}_T-\mathbf{u}_T]\odot M_p\|_2}
{\|\mathbf{u}_T\odot M_p\|_2},
\qquad \|\mathbf{u}_T\odot M_p\|_2>10^{-12}.
\label{eq:ood_terminal_error}
\end{equation}
This audit isolates morphology shift and does not establish real-rock, three-dimensional, parameter, or boundary-condition generalisation.

\subsection{Porous structure inverse design}
\label{sec:surrogate_design}

Inverse design is treated as deployment of the forward computational surrogate. A property target defines a search condition, candidate generators propose binary geometries, exact topology audits reject invalid structures, CT-PoreFlow screens connected candidates, and the LBM solver verifies the selected designs. Here, the conditional GAN and diffusion models are treated as alternative candidate-generation strategies.

The property vector characterising any structure $G$ is
\begin{equation}
\mathbf{c}(G)=[\phi,\phi_c,d,\log(1+k),\tau_h],
\label{eq:inverse_property_vector}
\end{equation}
and a design target specifies a desired vector $\mathbf{c}^{\dagger}$ in the same space.
During screening, morphological components ($\phi,\phi_c,d$) are calculated exactly
from the candidate mask, while terminal $k$ and $\tau_h$ are obtained from the
surrogate; this mixed predicted estimate is denoted $\widehat{\mathbf{c}}(G)$.
Candidate--target mismatch is
\begin{equation}
D(G,\mathbf{c}^{\dagger})=
\left\|
\mathbf{w}\odot
\frac{\widehat{\mathbf{c}}(G)-\mathbf{c}^{\dagger}}
{\mathbf{s}_{\mathrm{tr}}}
\right\|_2,
\qquad
\mathbf{w}=[1,0.7,0.7,2,1],
\label{eq:inverse_distance}
\end{equation}
where $\mathbf{s}_{\mathrm{tr}}$ contains training-split standard deviations. Eighteen targets are constructed from three training porosity quartiles, three local permeability quartiles, and two local tortuosity quartiles without test labels. Success@20 requires $|\Delta\phi|\leq0.03$, permeability error at most 20\%, and
tortuosity error at most 10\%. For the generative-design experiment, QC-conditional Success@20 is
calculated over candidates that pass the original LBM quality-control
criteria. Candidates that fail to converge or fail numerical quality
control are excluded from this conditional rate but remain represented
in the reported QC-pass denominator.

The finite-library comparison gives each method the same 764 unseen structures. Random-QSGS draws 50 candidates per target, morphology-only retrieval uses the first three distance terms, learned screens use predicted terminal fields, and an LBM oracle uses test labels only as a nondeployable attainable bound.

For generative search, the conditional GAN and conditional diffusion model are trained on the same 6078 training geometries and standardized property conditions. Both convert their continuous outputs to binary masks by zero thresholding; no connectivity repair, QSGS projection, or morphology filter is applied.

Each generator produces 256 masks for each of the 18 targets (4608 masks per
generator), giving 9216 raw generated structures in total across the conditional
GAN and conditional diffusion model. Exact morphology and through-connectivity are evaluated for every mask, and diversity uses unique-mask rate plus 2048 seeded pairwise Hamming comparisons per target. CT-PoreFlow predicts terminal $k$ and $\tau_h$ only for connected candidates. Six lowest-distance candidates and six disjoint uniformly sampled controls per method and target are assigned to LBM verification when the valid pool permits. Nonconvergence remains a failure in the assigned denominator, and target errors are calculated only for runs that pass the original residual, permeability-change, mass-balance, and Mach-number criteria.

The separate non-generative closed loop uses a fresh seeded pool of 120 QSGS structures whose identifiers are absent from all learning splits. CT-PoreFlow ranks three representative targets, and one guided and one uniformly sampled candidate per target are rerun with the D2Q9-BGK solver. In every design experiment, the surrogate performs screening; the final engineering property label is supplied only by the D2Q9-BGK LBM solver.

\section{Results and Validation}
\label{sec:results}
\subsection{Prediction performance}

QSGS-Transient-7606 covers a broad range of transport responses within the fixed two-dimensional QSGS--LBM setting. A total of 1,000 structures were requested in each of eight prescribed porosity intervals, yielding 649, 957, 1,000, 1,000, 1,000, 1,000, 1,000, and 1,000 accepted geometries, respectively. All 394 exhausted generation tasks occurred in the two lowest-porosity intervals and are retained in the audit manifests. Across the 7,606 accepted geometries, lattice permeability ranges from 0.0125 to 5.1442, with a median of 0.3552, while velocity-based hydraulic tortuosity ranges from 1.0366 to 2.2727, with a median of 1.2679.

\begin{figure}[pos=htbp]
\centering
\includegraphics[width=\textwidth,height=0.84\textheight,keepaspectratio]{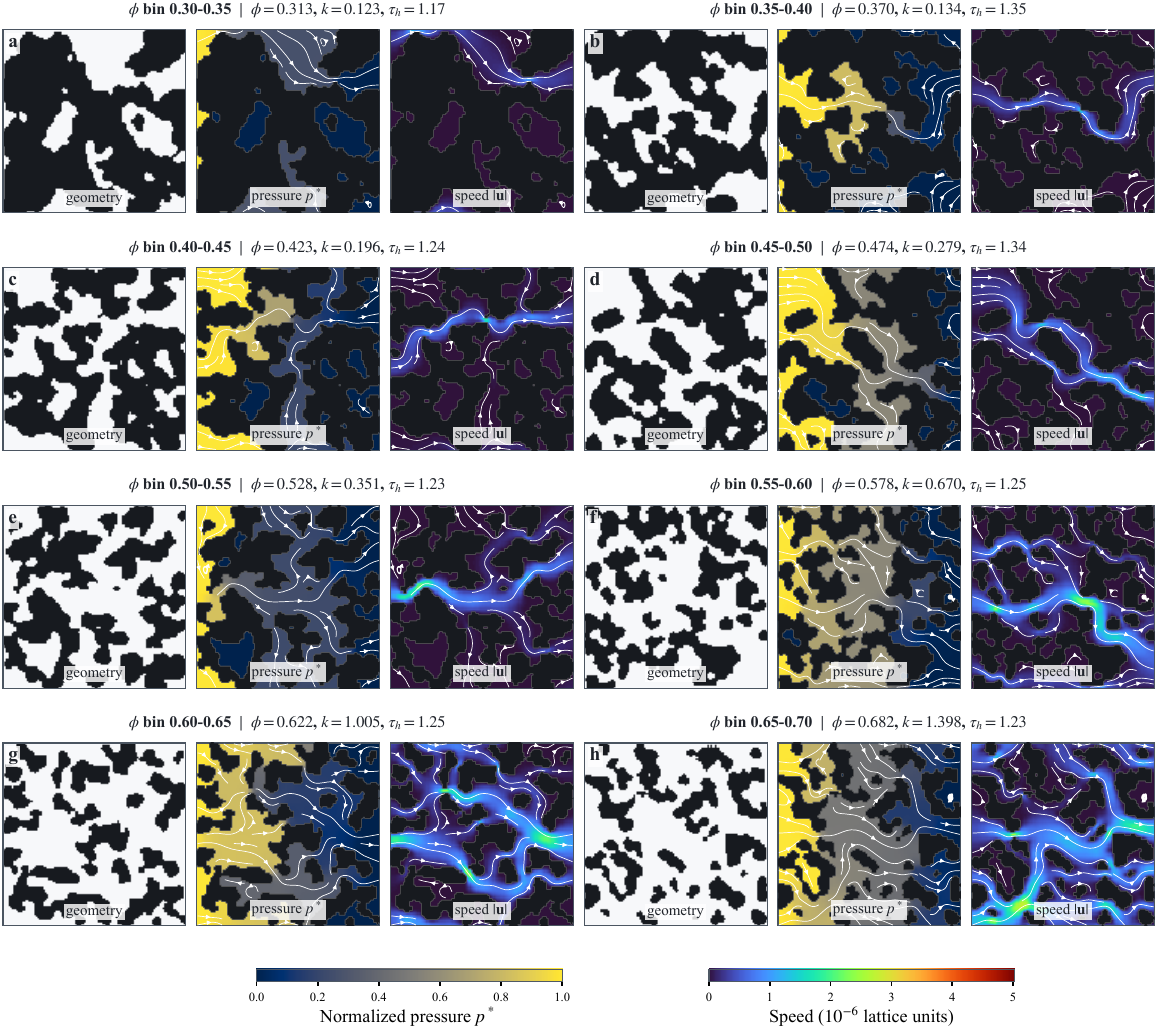}
\caption{Representative terminal LBM states across the prescribed porosity intervals.}
\label{fig:lbm_streamline_atlas}
\end{figure}

Fig.~\ref{fig:lbm_streamline_atlas} presents representative terminal LBM states across the eight porosity intervals. Within each interval, the displayed held-out geometry was selected near the multivariate median of total porosity, connected porosity, logarithmic permeability, and hydraulic tortuosity. The common pressure and velocity scales reveal that geometries with comparable total porosity can distribute flow through substantially different connected throats, resulting in distinct permeability and tortuosity. Thus, total porosity alone does not uniquely determine pore-scale transport within the present numerical regime.

\FloatBarrier

The observed within-interval variability further confirms that similar void fractions may correspond to different hydraulic-conductance and flow-path-complexity regimes. All 7,606 retained simulations satisfy the prescribed velocity-residual, permeability-change, mass-balance, and Mach-number criteria, and only one accepted realisation reaches the maximum limit of 50,000 iterations. These results characterise the numerical diversity and quality of the benchmark, but should not be interpreted as a constitutive relation beyond the fixed two-dimensional QSGS--LBM configuration.

The overall predictive comparison was conducted on the same 764 unseen geometries and all 30 stored time states for each model, resulting in 22,920 geometry--time records per model. Table~\ref{tab:main_results} reports the aggregate field errors, terminal engineering-property errors, mass-balance diagnostics, and model-only inference latency. Time-U-Net serves as the principal controlled comparator, while Time-FNO
and DeepONet are included as broader architectural references.

CT-PoreFlow achieved a velocity relative $L_2$ error of 0.2248, a normalised-pressure MAE of 0.01430, and a lattice-speed MAE of $3.965\times10^{-8}$. The corresponding Time-U-Net values were 0.2439, 0.01647, and $4.406\times10^{-8}$, representing reductions of 7.85\%, 13.19\%, and 10.00\%, respectively. CT-PoreFlow also reduced the terminal permeability error from 14.99\% to 12.81\%, the hydraulic-tortuosity error from 7.81\% to 5.10\%, and the mass-balance error from 0.0974 to 0.0839.

All six geometry-paired field and transport comparisons favoured CT-PoreFlow after Holm correction. For velocity prediction, the adjusted $p$-value was $9.66\times10^{-46}$, with a matched rank-biserial effect size of 0.598. DeepONet retained the lowest model-only inference latency at 0.54~ms frame$^{-1}$, compared with 1.32~ms frame$^{-1}$ for CT-PoreFlow. The results therefore support improved aggregate field and transport accuracy relative to Time-U-Net under the principal controlled comparison, rather than universal superiority over every architecture or evaluation metric.

\begin{table}[!ht]
\centering
\footnotesize
\setlength{\tabcolsep}{2.2pt}
\caption{Overall held-out field and engineering transport accuracy. Field metrics use 22,920 geometry--time records, whereas transport metrics use the terminal states of 764 unseen geometries.}
\label{tab:main_results}
\begin{tabularx}{\textwidth}{>{\raggedright\arraybackslash}Xrrrrrrr}
\hline
Method & Rel. $L_2$ & MAE $p^*$ & Speed MAE & $k$ error & $\tau_h$ error & Mass balance & ms/frame \\
\hline
Time-U-Net & 0.2439 & 0.0165 & $4.41\times10^{-8}$ & 14.99\% & 7.81\% & 0.0974 & 1.01 \\
Time-FNO & 0.4092 & 0.0304 & $7.39\times10^{-8}$ & 46.97\% & 17.09\% & 0.4558 & 0.79 \\
DeepONet & 0.8494 & 0.0833 & $1.98\times10^{-7}$ & 109.82\% & 148.25\% & 0.8935 & 0.54 \\
CT-PoreFlow & \textbf{0.2248} & \textbf{0.0143} & $\mathbf{3.97\times10^{-8}}$ & \textbf{12.81\%} & \textbf{5.10\%} & \textbf{0.0839} & 1.32 \\
\hline
\end{tabularx}
\end{table}

\subsection{Transient and engineering transport accuracy}

\begin{figure}[pos=htbp]
\centering
\includegraphics[width=\textwidth]{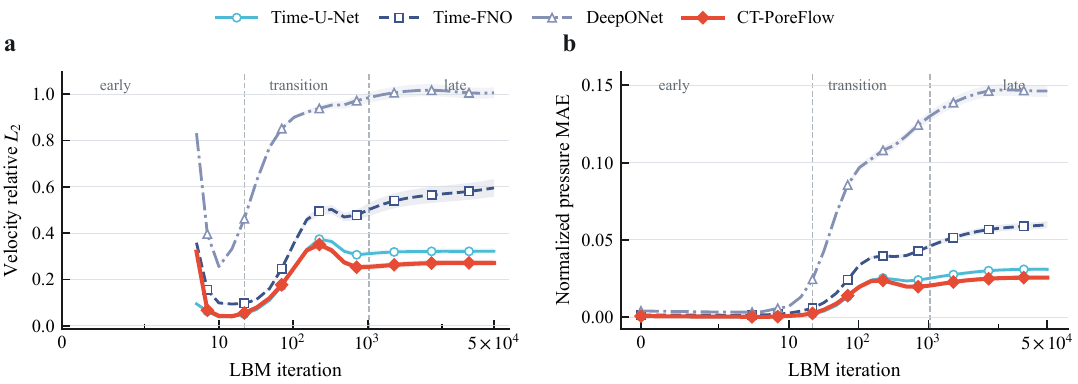}
\caption{Time-resolved velocity relative $L_2$ and normalised-pressure MAE on 764 held-out geometries.}
\label{fig:error_time}
\end{figure}

The time-resolved errors in Fig.~\ref{fig:error_time} reveal three characteristic stages in the development of pore-scale flow. At early times, the velocity magnitude remains close to zero, so the denominator of the relative $L_2$ metric becomes highly sensitive to small absolute deviations. Relative velocity error should therefore be interpreted together with the absolute speed error in this regime. During the transitional stage, the pressure field is established while the velocity field reorganises along the connected pore network, producing the most pronounced redistribution of active flow pathways. At late times, transport becomes localised within persistent connected throats and the prediction errors approach their terminal values. CT-PoreFlow exhibits lower aggregate errors than Time-U-Net across these stages, although it does not outperform the baseline at every individual stored time.

\FloatBarrier

\begin{figure}[pos=htbp]
\centering
\includegraphics[width=\textwidth]{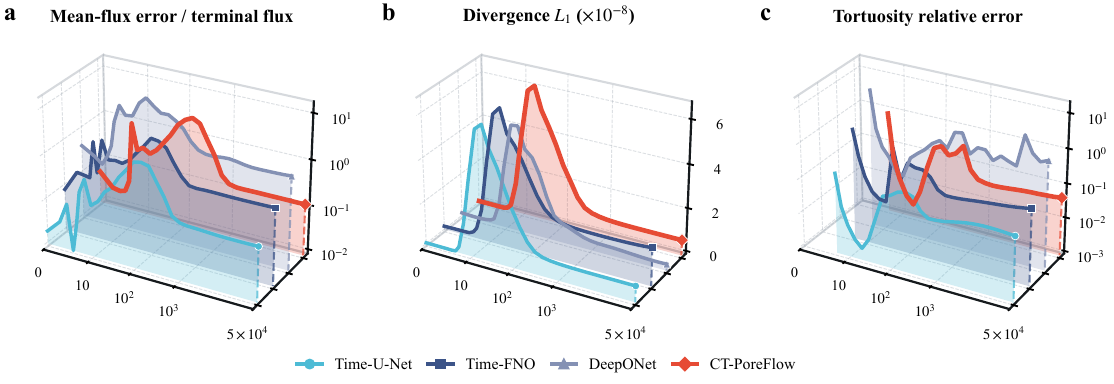}
\caption{Waterfall physical diagnostics: (a) terminal-flux-normalised mean-flux error, (b) interior divergence, and (c) hydraulic-tortuosity relative error.}
\label{fig:physical_dynamics}
\end{figure}

The predicted velocity and pressure fields are treated as complete numerical states from which flux, permeability, hydraulic tortuosity, divergence, and mass balance are recomputed. Permeability is therefore inferred from the reconstructed flow field rather than generated by an independent scalar-output head. Fig.~\ref{fig:physical_dynamics} tracks the evolution of these physical diagnostics. At the terminal state, CT-PoreFlow achieves a permeability error of 12.81\%, a hydraulic-tortuosity error of 5.10\%, and a mass-balance error of 0.0839, compared with 14.99\%, 7.81\%, and 0.0974 for Time-U-Net, respectively. The complete four-model comparison is summarised together with the field-level metrics in Table~\ref{tab:main_results}.

Exact pore masking enforces zero velocity within solid nodes for all models. However, the remaining divergence and inlet--outlet mass-balance errors demonstrate that geometric masking alone does not guarantee satisfaction of the continuity constraint. 

\FloatBarrier

\begin{figure}[pos=htbp]
\centering
\includegraphics[width=\textwidth]{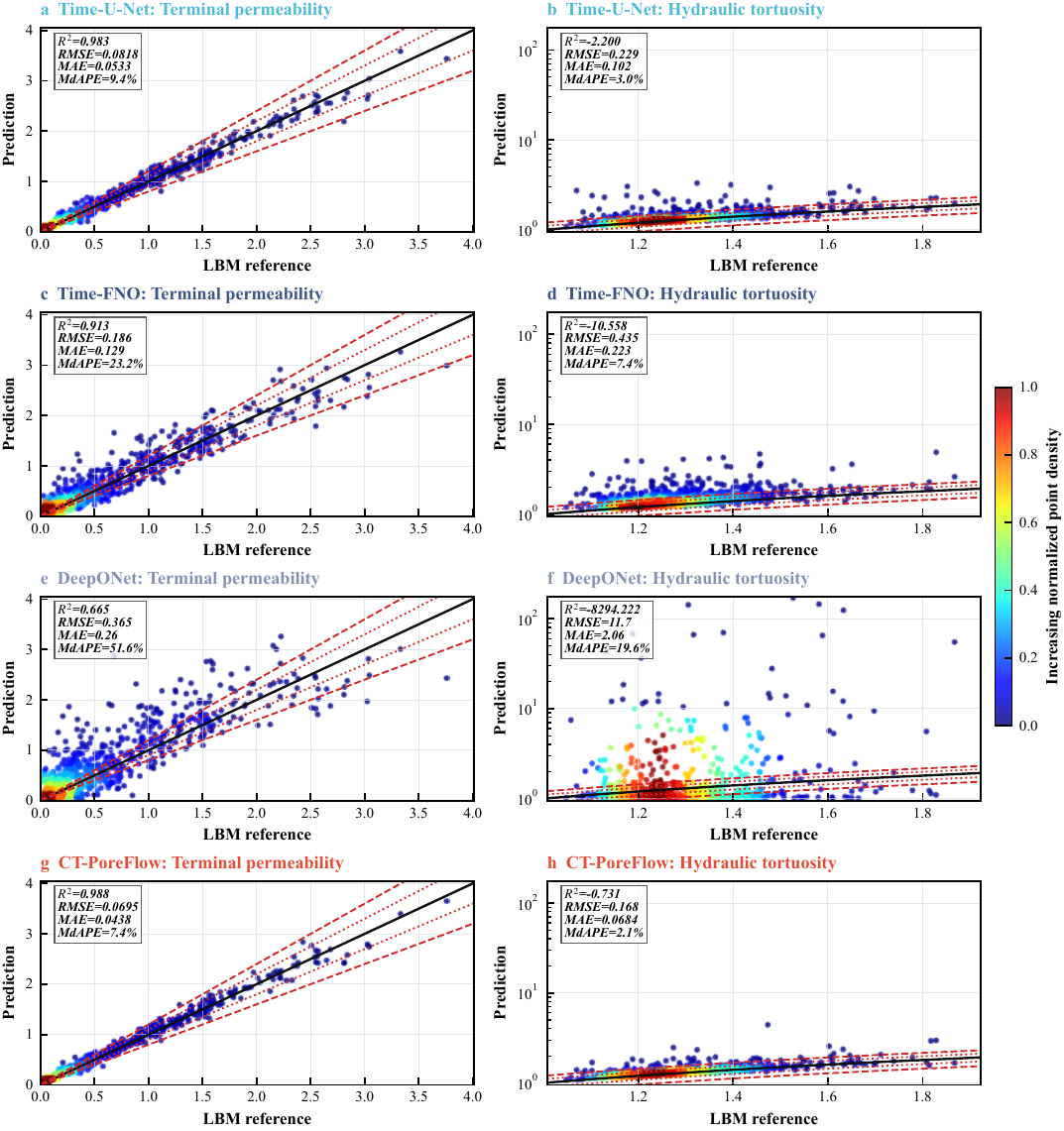}
\caption{Terminal permeability and hydraulic-tortuosity calibration for 764 paired held-out geometries.}
\label{fig:physical_calibration}
\end{figure}

The terminal calibration results in Fig.~\ref{fig:physical_calibration} further show that engineering-property errors are strongly property dependent. Permeability is well calibrated for the two most accurate field surrogates, with $R^2=0.988$ for CT-PoreFlow and $R^2=0.983$ for Time-U-Net. Hydraulic tortuosity is considerably more difficult to recover. Long-tail prediction failures produce negative coefficients of determination for all four models: $R^2=-0.731$ for CT-PoreFlow, $-2.200$ for Time-U-Net, $-10.558$ for Time-FNO, and $-8294.222$ for DeepONet.

For CT-PoreFlow, the negative tortuosity $R^2$ coexists with a low median absolute percentage error of 2.1\%. This combination indicates that the model provides accurate tortuosity estimates for most geometries but exhibits substantial errors for a small number of extreme cases. Because $R^2$ is highly sensitive to these long-tail failures, whereas the median percentage error is not, both metrics are required to characterise performance adequately. Such adverse cases are important in pore-scale transport because local errors in dominant path length or flow-channel organisation may remain concealed by aggregate field metrics or median statistics. The observed calibration spread therefore supports the use of tortuosity
predictions for relative candidate ranking rather than direct property assignment.

\FloatBarrier

These results demonstrate that field-level and transport-level model rankings are
not interchangeable. The Direct transient decoder, evaluated before late-time
flux calibration, yields the lowest aggregate field error, whereas CT-PoreFlow
provides the lowest terminal transport error after calibration. Minimising the
average discrepancy in the reconstructed velocity and pressure fields therefore
does not necessarily minimise the error in Darcy-scale transport quantities.
CT-PoreFlow is therefore interpreted as a transport-oriented Pareto solution
rather than the optimum under every individual metric.

Within the benchmark, total porosity does not uniquely determine permeability, as evidenced by the substantial within-interval spread in transport properties. Connected porosity is more directly related to hydraulic conductance, exhibiting a Spearman rank correlation of $\rho=0.80$ with lattice permeability. In contrast, the dead-pore ratio, $1-\phi_c/\phi$, has a negative correlation of $\rho=-0.72$, consistent with the reduction in hydraulically active pore volume as the nonspanning pore fraction increases. These relationships distinguish total void fraction from effective transport support, although they should not be interpreted as universal constitutive laws beyond the present QSGS--LBM configuration.

Hydraulic tortuosity captures a distinct aspect of pore-network organisation. Its global rank correlation with connected porosity is approximately zero, indicating that a large connected pore volume does not necessarily correspond to a short or direct transport pathway. The 7,606 accepted structures therefore span multiple regimes, including low-conductance/long-path, high-conductance/long-path, and high-conductance/short-path configurations. Permeability and hydraulic tortuosity consequently provide complementary descriptions of hydraulic conductance and path complexity. This separation also defines the appropriate property space for the subsequent inverse-design analysis. Retaining both permeability and hydraulic tortuosity prevents structures with comparable conductance but substantially different internal pathway organisation from being treated as equivalent.

\subsection{Model robustness and generalisation}

The principal benchmark evaluates model performance on unseen QSGS geometries at the time states represented in the training schedule. To examine robustness beyond this setting, three complementary analyses were conducted: component ablation, prediction at excluded time indices, and transfer to geometries outside the QSGS training distribution. A spatial failure analysis was further performed to identify the local flow features responsible for the aggregate errors.

The ablation experiments isolate the contributions of the geometry representation, global information exchange, temporal conditioning, decoder formulation, and transport-oriented loss. Removing the directed geodesic channels, $D_{\mathrm{in}}$ and $D_{\mathrm{out}}$, while retaining all other CT-PoreFlow components decreases the aggregate velocity error by 0.79\%. Thus, although these channels encode inlet- and outlet-referenced transport positions that cannot be represented by the Euclidean distance transform alone, they do not provide a uniform improvement in aggregate field accuracy under the present configuration.

By contrast, global information exchange has a pronounced effect. Removing the FNOBlock increases the velocity error by 26.78\%, representing the largest major degradation among the component-removal experiments. This result supports the use of compressed spectral mixing for communicating information across spatially separated regions of the connected pore network, although it does not imply that all topology-related inputs are equally beneficial.

The temporal and transport formulations introduce a different set of trade-offs. Before flux calibration, the Direct transient decoder reduces the velocity error by 12.19\% and the terminal flux error by 65.68\% relative to the Terminal-anchored steady-residual decoder. Replacing logarithmic time conditioning with a linear time coordinate increases the velocity error by 14.40\%, while using the field loss alone increases it by 11.71\%. Late-time flux calibration subsequently trades a modest increase in aggregate field error for improved terminal transport recovery.

\begin{table}[t]
\centering
\caption{Comparison of CT-PoreFlow decoder and flux-calibration
configurations, with Time-U-Net included as a reference baseline.}
\label{tab:ablation_results}
\small
\begin{tabularx}{\textwidth}{>{\raggedright\arraybackslash}Xrrrr}
\hline
Prediction and loss configuration & Rel. $L_2$ velocity & MAE $p^*$ & Flux error & Parameters (M) \\
\hline
Terminal-anchored steady-residual decoder & 0.2435 & 0.0161 & 54.76\% & 12.12 \\
Direct transient decoder & \textbf{0.2138} & \textbf{0.0138} & 18.79\% & 10.21 \\
CT-PoreFlow & 0.2248 & 0.0143 & \textbf{12.81\%} & 10.21 \\
Time-U-Net & 0.2439 & 0.0165 & 14.99\% & 8.70 \\
\hline
\end{tabularx}
\end{table}

The unseen-time audit evaluates 10 stored time indices that were excluded from both training and validation. On these LBM states, CT-PoreFlow obtained a velocity relative $L_2$ error of 0.3267, a pressure MAE of 0.01977, and a speed MAE of $6.40\times10^{-8}$. The corresponding Time-U-Net values were 0.3224, 0.01950, and $6.06\times10^{-8}$, respectively. CT-PoreFlow therefore did not outperform the baseline in this experiment.

The query-time formulation permits direct evaluation at an arbitrary requested time without recursively advancing through previous predicted states. However, this computational capability should be distinguished from interpolation accuracy. The excluded-time results do not demonstrate superior interpolation between the stored training states, while off-grid animation queries lack direct LBM references and remain qualitative. Neither analysis provides evidence of temporal extrapolation beyond the simulated time range.

Morphological transfer was first evaluated using 24 rule-based geometries from four families absent from the QSGS training distribution: circular arrays, snowflake obstacles, tree-like obstacles, and staggered baffles. Each family contains two porosity levels and three seeded replicates at each level. The mean terminal velocity relative $L_2$ errors over the four families are 0.3084 for Time-FNO, 0.3183 for CT-PoreFlow, 0.3997 for Time-U-Net, and 0.8233 for DeepONet. Time-FNO therefore achieves the lowest overall error, while CT-PoreFlow remains approximately 20.4\% lower than Time-U-Net. The result supports competitive morphology transfer but not universal superiority of CT-PoreFlow across all out-of-distribution settings.

\begin{figure}[pos=htbp]
\centering
\includegraphics[width=\textwidth]{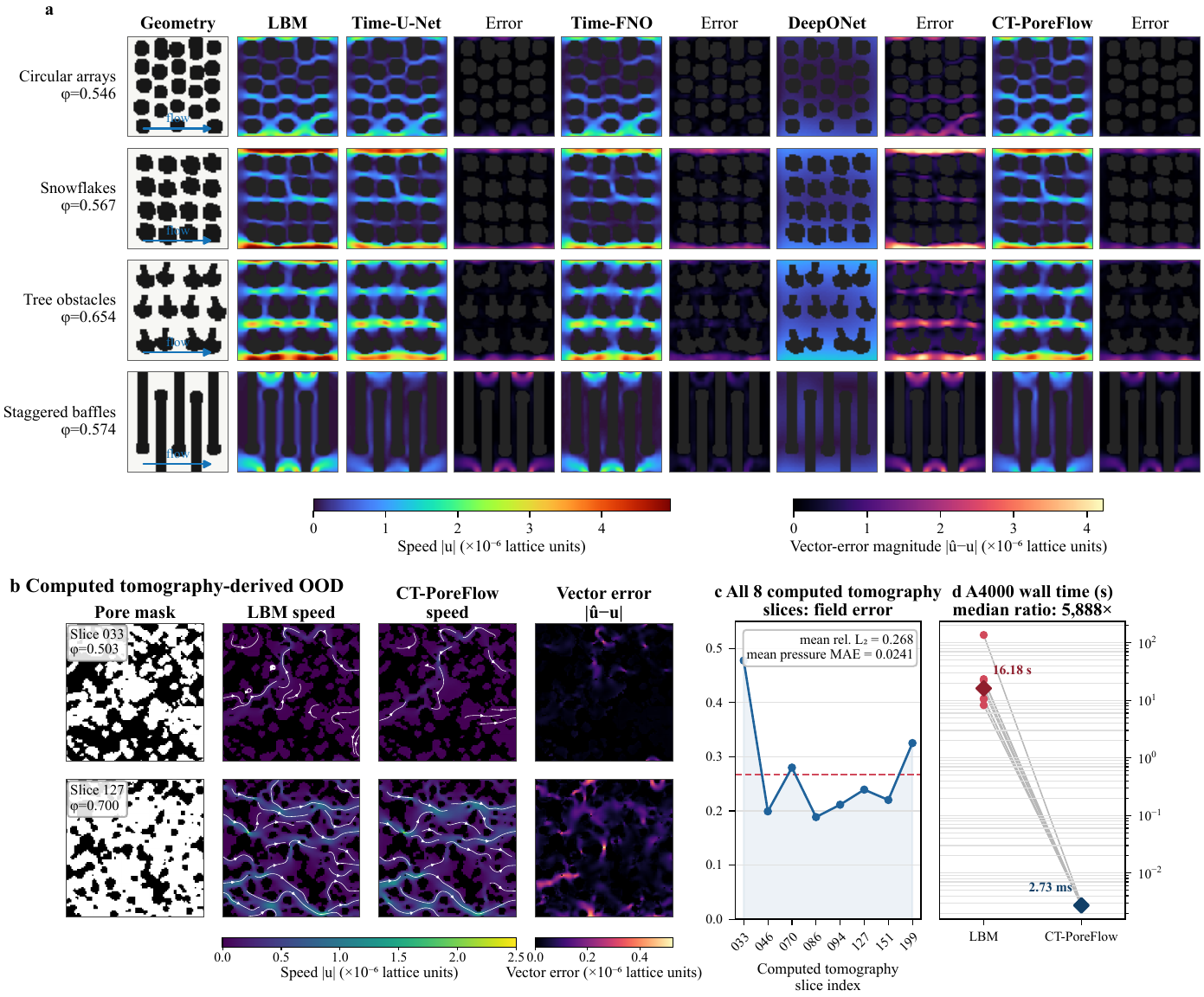}
\caption{Terminal morphology OOD validation. (a) Rule-based families absent from QSGS training. (b) Image-derived OOD examples from the minimum- and maximum-porosity computed tomography slices, selected before viewing prediction errors. (c) Terminal velocity relative $L_2$ for every QC-passing computed tomography slice. (d) Same-node RTX A4000 wall time for the LBM terminal solve and synchronized model-only inference.}
\label{fig:rule_ood_fields}
\end{figure}

The upper part of Fig.~\ref{fig:rule_ood_fields} compares the reconstructed terminal fields for the unfamiliar rule-based geometries. Under a common physical speed scale, Time-U-Net, Time-FNO, and CT-PoreFlow preserve the principal LBM transport channels, whereas DeepONet smooths or suppresses several constriction-driven jets. For CT-PoreFlow, the remaining discrepancies are spatially concentrated near narrow throats, obstacle tips, and inlet--outlet transitions rather than being uniformly distributed throughout the pore space. The lower part of the figure extends the analysis to image-derived CT masks using the same left-to-right pressure forcing and the frozen CT-PoreFlow checkpoint.

\FloatBarrier

All eight computed tomography slices contained a connected left-to-right pore pathway and satisfied the LBM convergence and mass-balance criteria. Without fine-tuning, the terminal velocity relative $L_2$ error ranges from 0.1886 to 0.4778, with a mean of 0.2677 and a median of 0.2298. The mean normalised-pressure MAE is 0.02410. Slice 033 exhibits the largest field error and the lowest connected pore fraction, 0.512, whereas the velocity errors for the remaining seven slices range from 0.1886 to 0.3253. This relationship is descriptive because all eight slices are correlated observations obtained from a single segmented volume.

On the same RTX A4000 node, the median terminal LBM solution time is 16.18~s per slice, compared with 2.73~ms for synchronized model-only inference, corresponding to a median ratio of $5{,}888\times$. The mean times are 30.54~s and 2.80~ms, respectively, because the least-connected slice reaches the 50,000-iteration limit. These values compare one terminal LBM simulation with one frozen-network evaluation and represent implementation-level timing measurements rather than modifications to the underlying physical model.

\begin{figure}[pos=htbp]
\centering
\includegraphics[width=\textwidth]{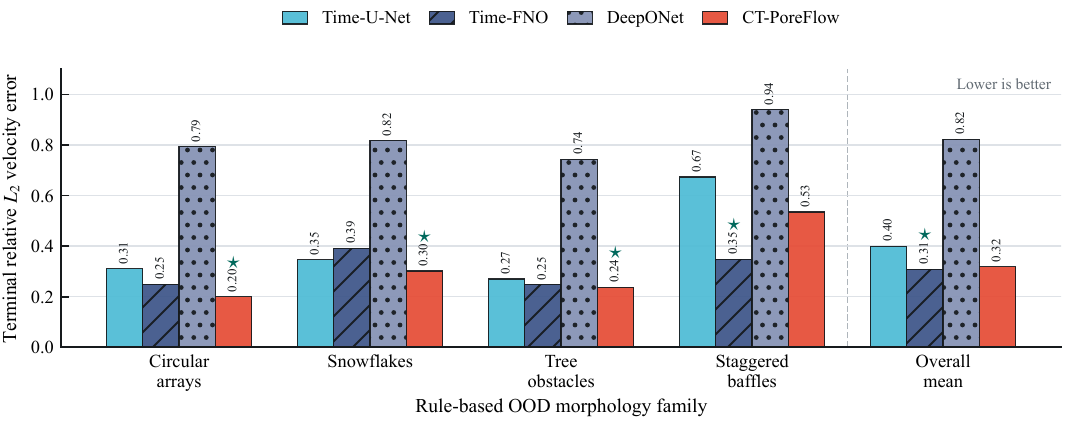}
\caption{Frozen-model performance on the rule-based morphology OOD set.}
\label{fig:rule_ood_models}
\end{figure}

Fig.~\ref{fig:rule_ood_models} resolves the overall result by morphology family. CT-PoreFlow achieves the lowest error for circular arrays, snowflake obstacles, and tree-like obstacles. Time-FNO performs substantially better for staggered baffles, with errors of 0.3476 and 0.5348 for Time-FNO and CT-PoreFlow, respectively. The family-dependent ranking indicates that transfer performance depends not only on porosity or connectivity, but also on the spatial organisation of the pore and obstacle networks.

\FloatBarrier

\begin{figure}[pos=htbp]
\centering
\includegraphics[width=\textwidth,height=0.72\textheight,keepaspectratio]{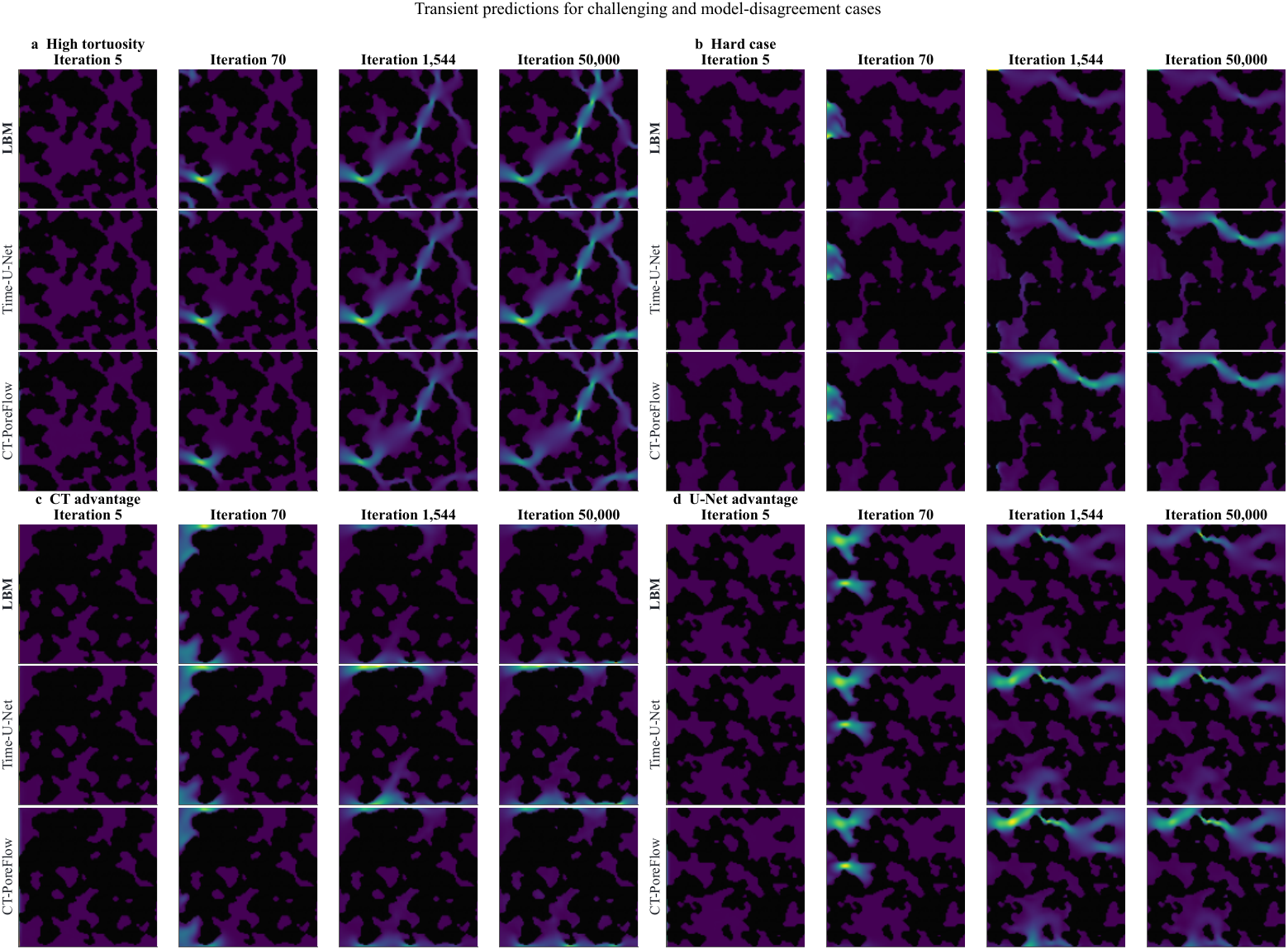}
\caption{Transient comparison for high tortuosity, the highest joint-error case, and the strongest CT-PoreFlow and Time-U-Net advantage cases.}
\label{fig:gallery_transient_challenges}
\end{figure}

To identify the local origins of the aggregate errors, Fig.~\ref{fig:gallery_transient_challenges} examines a high-tortuosity geometry, the geometry with the highest joint error, and the cases showing the strongest relative advantages of CT-PoreFlow and Time-U-Net. Each geometry is evaluated at four stored time states.

For the high-tortuosity geometry, both surrogates recover the dominant connected pathway as the flow develops, while the principal discrepancies occur near the upper inlet region and the downstream high-speed throat. In the highest joint-error case, both models identify the main vertical pathway, but errors in its width and velocity magnitude persist into the developed state. The two model-advantage cases further demonstrate that prediction differences are predominantly local: one model may reconstruct a boundary-driven jet or narrow constriction more accurately even when both models recover the same large-scale connected support.

The early-time fields also illustrate the limitations of relative error near rest. Weak motion initially develops near the driven boundaries before propagating through the spanning pore network. Small absolute differences can therefore generate large relative errors while the reference velocity norm remains small. At later times, prediction quality depends more directly on pathway localisation, including whether the dominant throat is reconstructed, whether spurious flow develops in dead-end pores, and whether a localised jet is broadened or suppressed.

For each geometry--time combination, the LBM, Time-U-Net, and CT-PoreFlow panels use a common velocity scale. The scale is allowed to vary between geometries and time states to make weak early-time motion visible; colour intensity should therefore not be compared quantitatively across the complete montage. The visual analysis identifies the spatial origin of the numerical errors but does not replace the tensor-based evaluation.

\FloatBarrier

The combined robustness analyses lead to three principal observations. First, independent query-time evaluation avoids recurrent error propagation \citep{shi2015convlstm,kumar2025adcrnn}, but does not by itself guarantee superior interpolation at unseen time indices. Second, architectural priors provide nonuniform benefits. Directed geodesic channels encode boundary-referenced graph positions that are absent from the Euclidean distance transform, whereas the FNOBlock supports long-range communication through compressed spectral mixing. Although related geodesic augmentation has improved transient reactive-transport prediction \citep{kim2026prtdeeponet}, the present momentum-relaxation problem does not show a uniform aggregate-field benefit from the directed channels. The FNOBlock, by contrast, provides the clearest architectural contribution.

Geometries with restricted connected support, substantial dead-pore volume, low permeability, or long transport paths can remain difficult even when their total porosity is not unusual. This behaviour is consistent with pore-scale interpretations of hydraulic tortuosity \citep{matyka2008tortuosity} and supports reporting connected porosity and dead-pore ratio alongside total porosity.

Finally, visual plausibility should not be used as a substitute for physical diagnostics. A single training-derived speed limit is applied across samples, time states, references, and predictions so that colour retains a consistent lattice-velocity interpretation. Exact masking eliminates velocity leakage into solid nodes, but it does not enforce divergence-free flow or inlet--outlet mass conservation. These quantities must therefore be evaluated explicitly from the numerical fields.

\section{Surrogate-Assisted Inverse Design and LBM Verification}
\label{sec:inverse}

Section~\ref{sec:results} established the field and transport accuracy of
CT-PoreFlow on unseen geometries. The following analyses evaluate its performance
in three increasingly demanding design settings: ranking within a held-out
library, screening a fresh QSGS pool, and property-conditioned generation followed
by high-fidelity verification.

\subsection{Surrogate-guided targeted search}
\label{sec:inverse_targeted_search}

\begin{figure}[pos=htbp]
\centering
\includegraphics[width=\textwidth]{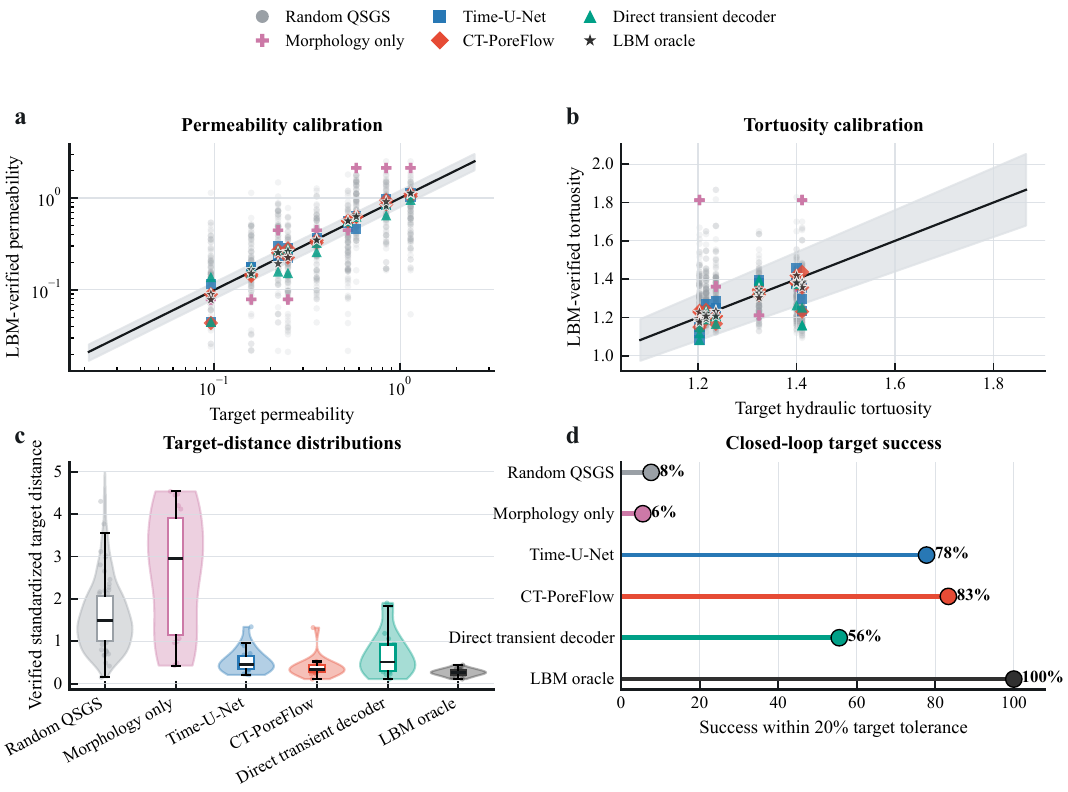}
\caption{Controlled inverse retrieval in the held-out LBM library. (a,b) Requested versus LBM-verified permeability and hydraulic tortuosity, with equality lines, tolerance bands, the full random-QSGS cloud, and deterministic selections from the learned and ablated methods. (c) Violin, box, and sample-point summaries of the complete target-distance distributions. (d) Success@20.}
\label{fig:inverse_calibration}
\end{figure}

The finite-library experiment provides a controlled assessment of surrogate-based candidate ranking rather than a complete inverse-design demonstration. As defined in Section~\ref{sec:surrogate_design}, 18 target property conditions were evaluated against a common library of 764 unseen geometries. The candidates selected by each method were subsequently assessed using their LBM-labelled properties. The resulting calibration and target-distance distributions are presented in Fig.~\ref{fig:inverse_calibration}, while the corresponding quantitative results are summarised in Table~\ref{tab:inverse_library}.

CT-PoreFlow achieved a mean LBM-verified target distance of 0.386, a permeability target error of 11.44\%, a hydraulic-tortuosity target error of 2.80\%, and a Success@20 rate of 83.33\%. Time-U-Net obtained a mean distance of 0.537 and a Success@20 rate of 77.78\%, whereas the Direct transient decoder obtained 0.700 and 55.56\%, respectively. By comparison, the random-QSGS reference, based on 50 independent draws per target, produced a mean distance of 1.618 and a Success@20 rate of 7.67\%.

An important contrast is observed between field-level and design-level performance. Although the Direct transient decoder yields a lower aggregate field error than CT-PoreFlow in Section~\ref{sec:results}, its candidate-ranking accuracy is substantially poorer. This result demonstrates that reducing pixel-averaged field error does not necessarily improve the ranking of geometries according to engineering transport objectives. The LBM-oracle, which uses the true test labels, provides a nondeployable upper bound with a mean distance of 0.262. The remaining difference between CT-PoreFlow ($D=0.386$) and the oracle ($D=0.262$) reflects both surrogate-ranking error and the finite coverage of the candidate library.

\FloatBarrier

\begin{table}[t]
\centering
\footnotesize
\caption{Controlled surrogate-ranking benchmark over 18 property targets and 764 unseen geometries.}
\label{tab:inverse_library}
\begin{tabularx}{\textwidth}{>{\raggedright\arraybackslash}Xrrrrr}
\hline
Method & Selections & Mean $D$ & $k$ target error & $\tau_h$ target error & Success@20 (\%) \\
\hline
Random-QSGS & 900 & 1.618 & 64.79\% & 8.84\% & 7.67 \\
Morphology-only & 18 & 2.626 & 87.22\% & 16.82\% & 5.56 \\
Time-U-Net & 18 & 0.537 & 13.04\% & 4.33\% & 77.78 \\
CT-PoreFlow & 18 & \textbf{0.386} & \textbf{11.44\%} & \textbf{2.80\%} & \textbf{83.33} \\
Direct transient decoder & 18 & 0.700 & 18.03\% & 6.04\% & 55.56 \\
LBM-oracle & 18 & 0.262 & 7.30\% & 1.54\% & 100.00 \\
\hline
\end{tabularx}
\vspace{2pt}

\begin{minipage}{\textwidth}
\footnotesize\emph{Note:} Random-QSGS is a random-search reference based on 50 independent draws per target. The LBM-oracle is a nondeployable attainable bound constructed from the true test labels.
\end{minipage}
\end{table}

The finite-library experiment evaluates ranking performance within a fixed candidate set. A more stringent physical round-trip test was therefore performed using a newly generated pool of 120 QSGS structures whose identifiers were absent from the training, validation, and test splits. Three representative design objectives were considered: high permeability, low hydraulic tortuosity, and a joint-transport target. For each objective, CT-PoreFlow ranked the fresh candidates and selected one guided structure, while one additional structure was sampled as a random control. All six candidates were then recomputed using the D2Q9-BGK solver.

Across the three paired targets, surrogate-guided selection reduced the mean LBM-verified target distance from 1.494 to 0.335, corresponding to a descriptive reduction of 77.59\%. Because the experiment contains only $n=3$ paired targets, no formal significance test is reported. All six LBM simulations satisfied the original convergence, velocity-residual, permeability-change, mass-balance, and Mach-number acceptance criteria.

The benefit of guided screening varied across the three objectives. For the high-permeability target, the verified distance decreased from 2.054 for the random candidate to 0.390 for the guided candidate. For the joint-transport target, the corresponding reduction was from 1.391 to 0.277. For the low-$\tau_h$ target, guided screening reduced the verified distance from 1.038 to 0.337; however, the resulting permeability remained 17.31\% above the requested value. This residual mismatch illustrates the coupled nature of the transport objectives, because permeability and tortuosity depend on the same connected pathways and cannot necessarily be optimised independently.

\begin{figure}[pos=htbp]
\centering
\includegraphics[width=\textwidth,height=0.80\textheight,keepaspectratio]{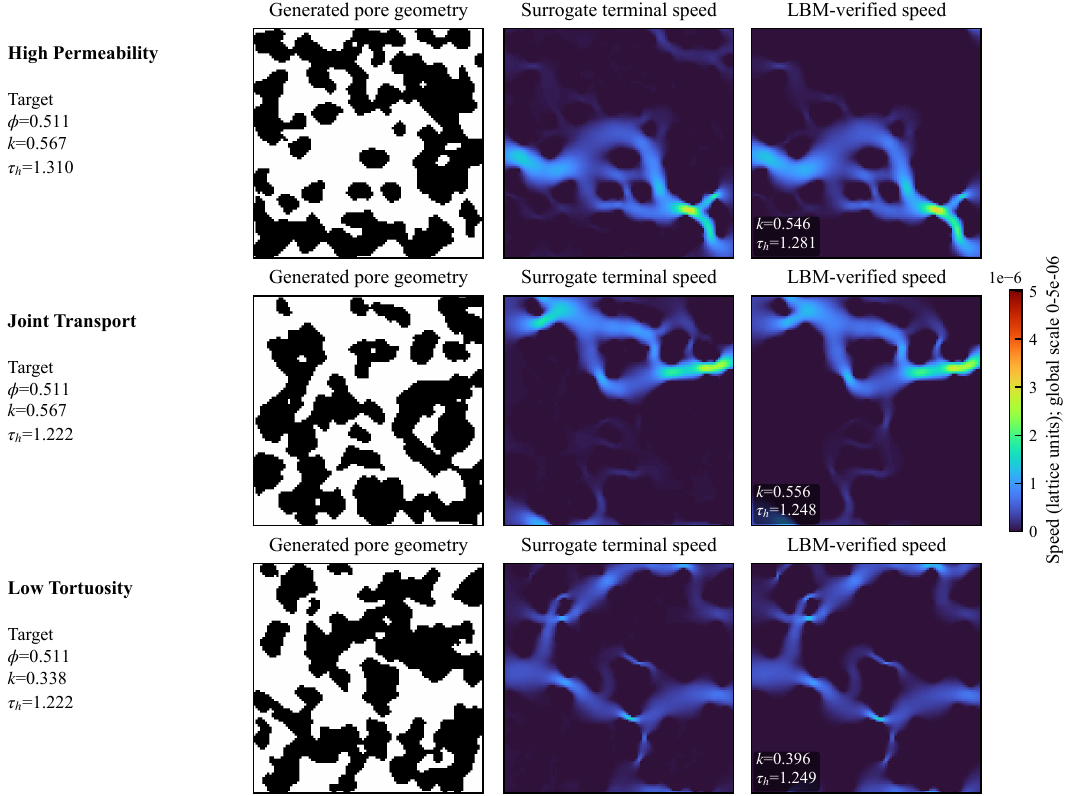}
\caption{LBM-verified fresh-QSGS guided designs.}
\label{fig:inverse_cases}
\end{figure}

Fig.~\ref{fig:inverse_cases} presents the corresponding pore geometries, surrogate-predicted terminal fields, and LBM-verified responses. The agreement between the selected candidates and their high-fidelity recomputations illustrates the potential value of the surrogate for reducing the
candidate-search space

\FloatBarrier

\subsection{Property-conditioned generative design and LBM verification}
\label{sec:property_conditioned_design}
\label{sec:generated_lbm_verification}

The property-conditioned design experiment evaluates whether prescribed engineering targets can be translated into candidate pore populations that simultaneously exhibit hydraulic validity, target fidelity, and structural diversity. Under the protocol defined in Section~\ref{sec:surrogate_design}, the conditional GAN and conditional diffusion model each generated 4,608 binary masks across the same 18 target conditions, resulting in 9,216 candidates in total. Candidate quality was first assessed through exact topology analysis and surrogate-based transport screening, after which selected candidates were returned to the D2Q9-BGK solver for LBM verification.

\begin{figure}[pos=htbp]
\centering
\includegraphics[width=\textwidth]{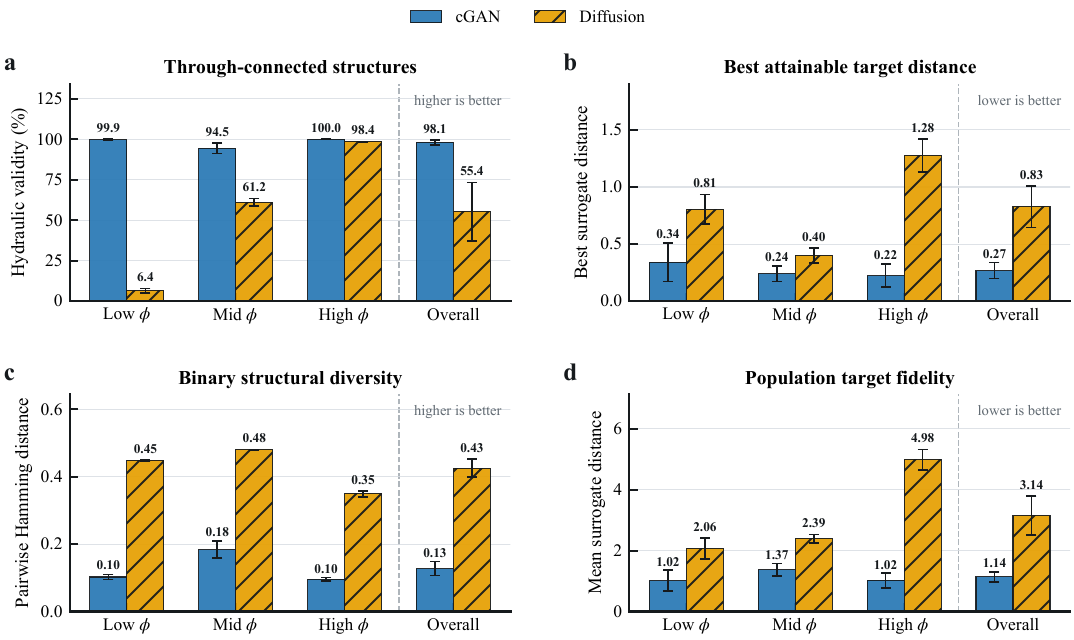}
\caption{Property-conditioned candidate quality across low-, middle-, and high-porosity specifications. Grouped bars compare the conditional GAN and conditional diffusion model in terms of (a) through-connectivity, (b) best attainable surrogate target distance, (c) pairwise binary Hamming diversity, and (d) mean surrogate target distance.}
\label{fig:generative_quality}
\end{figure}

The two generators exhibited markedly different validity--diversity--fidelity characteristics. The conditional GAN achieved a mean through-connectivity of 98.11\%, compared with 55.36\% for conditional diffusion. As shown in Fig.~\ref{fig:generative_quality}, the reduced connectivity of the diffusion model was concentrated primarily in the low-porosity target group rather than being uniform across all target conditions.

At the surrogate-screening level, the conditional GAN obtained a mean proxy distance of 1.136 and a best proxy distance of 0.268, whereas conditional diffusion obtained 3.144 and 0.827, respectively. Structural diversity showed the opposite trend. The mean pairwise Hamming distance was 0.127 for the conditional GAN and 0.425 for conditional diffusion, while both generators produced 100\% unique thresholded masks. Thus, diffusion explored a broader binary morphology space, whereas the conditional GAN generated a narrower but more hydraulically connected and target-compatible candidate population under the present implementation and sampling budget. Generation and surrogate screening required 54.75~ms per conditional-GAN candidate and 193.87~ms per conditional-diffusion candidate.

\FloatBarrier

These screening-level metrics characterise the quality of the generated candidate populations but do not establish physical design success. In particular, uniqueness and Hamming diversity quantify morphological variation without guaranteeing inlet--outlet connectivity, numerical convergence, or agreement with the requested transport properties.

The LBM solver accepted 101 of 108 surrogate-guided conditional-GAN candidates, 101 of 108 randomly selected conditional-GAN candidates, 108 of 108 guided conditional-diffusion candidates, and 104 of 105 random conditional-diffusion candidates. Acceptance required satisfaction of the original convergence, velocity-residual, permeability-change, mass-balance, and Mach-number criteria. Rejected simulations remain included in the assigned-candidate denominator reported by the QC-pass metric, but are excluded from the property-error and Success@20 summaries. The corresponding mean LBM runtimes were 19.70, 20.55, 12.92, and 13.49~s, respectively.

\begin{table}[t]
\centering
\footnotesize
\caption{LBM verification of surrogate-selected and random generated candidates. Mean $D$, property errors, and Success@20 are candidate-level summaries over QC-passing runs. QC pass reports accepted over assigned candidates; rejected runs remain in this QC denominator but are excluded from the property-error and Success@20 summaries.}
\label{tab:generative_lbm}
\begin{adjustbox}{width=\textwidth}
\begin{tabular}{llcrrrr}
\hline
Generator & Selection & QC pass & Mean $D$ & $k$ error (\%) & $\tau_h$ error (\%) & Success@20 (\%) \\
\hline
Conditional GAN & Guided & 101/108 & \textbf{0.481} & \textbf{19.16} & \textbf{3.42} & \textbf{72.28} \\
Conditional GAN & Random & 101/108 & 1.160 & 58.20 & 6.13 & 22.77 \\
Conditional diffusion & Guided & 108/108 & 1.202 & 32.28 & 5.69 & 10.19 \\
Conditional diffusion & Random & 104/105 & 3.233 & 190.31 & 10.33 & 0.96 \\
\hline
\end{tabular}
\end{adjustbox}
\end{table}

Table~\ref{tab:generative_lbm} reports candidate-level statistics over QC-passing simulations. For paired statistical inference, the results were first aggregated within each of the 18 target conditions. For the conditional GAN, surrogate-guided selection reduced the target-level mean verified distance from 1.168 for random selection to 0.508, corresponding to a paired reduction of 0.659. The 95\% bootstrap confidence interval was $[0.549,\,0.765]$, and the exact sign-flip test gave $p=7.63\times10^{-6}$. For conditional diffusion, the target-level mean distance decreased from 3.180 to 1.202, giving a paired reduction of 1.978 with a 95\% bootstrap confidence interval of $[1.375,\,2.619]$ and an exact sign-flip value of $p=1.53\times10^{-5}$. All 18 target conditions retained at least one QC-passing candidate under both guided and random selection for each generator.

The candidate-level results provide the same overall ranking. Surrogate-guided conditional-GAN selection achieved a mean verified distance of 0.481 and a Success@20 rate of 72.28\%, compared with 1.160 and 22.77\% for random GAN selection. Guided conditional diffusion reduced the mean verified distance from 3.233 to 1.202 and increased Success@20 from 0.96\% to 10.19\%. These results demonstrate that surrogate screening improves candidate selection for both generators, although the final performance remains strongly dependent on the validity and target compatibility of the underlying generated population.

\begin{figure}[pos=htbp]
\centering
\includegraphics[width=\textwidth]{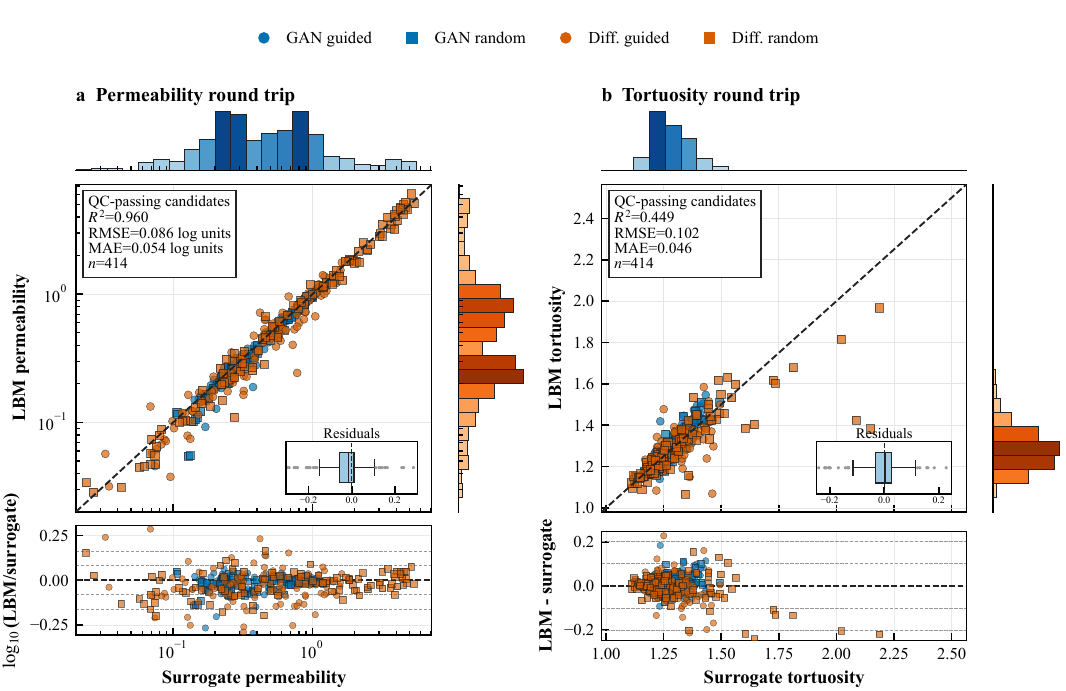}
\caption{Surrogate-to-LBM round-trip calibration for the 414 QC-passing generated candidates.}
\label{fig:generative_lbm}
\end{figure}

The independent surrogate-to-LBM comparison in Fig.~\ref{fig:generative_lbm} includes all 414 QC-passing generated candidates. Permeability exhibits close agreement between the surrogate and LBM results, with $R^2=0.960$, RMSE $=0.086$ log units, and MAE $=0.054$ log units. Hydraulic tortuosity is less tightly calibrated, with $R^2=0.449$, RMSE $=0.102$, and MAE $=0.046$. This difference is consistent with the greater sensitivity of tortuosity to local pathway organisation and long-tail geometric cases.

\FloatBarrier

\begin{figure}[pos=htbp]
\centering
\includegraphics[width=\textwidth,height=0.80\textheight,keepaspectratio]{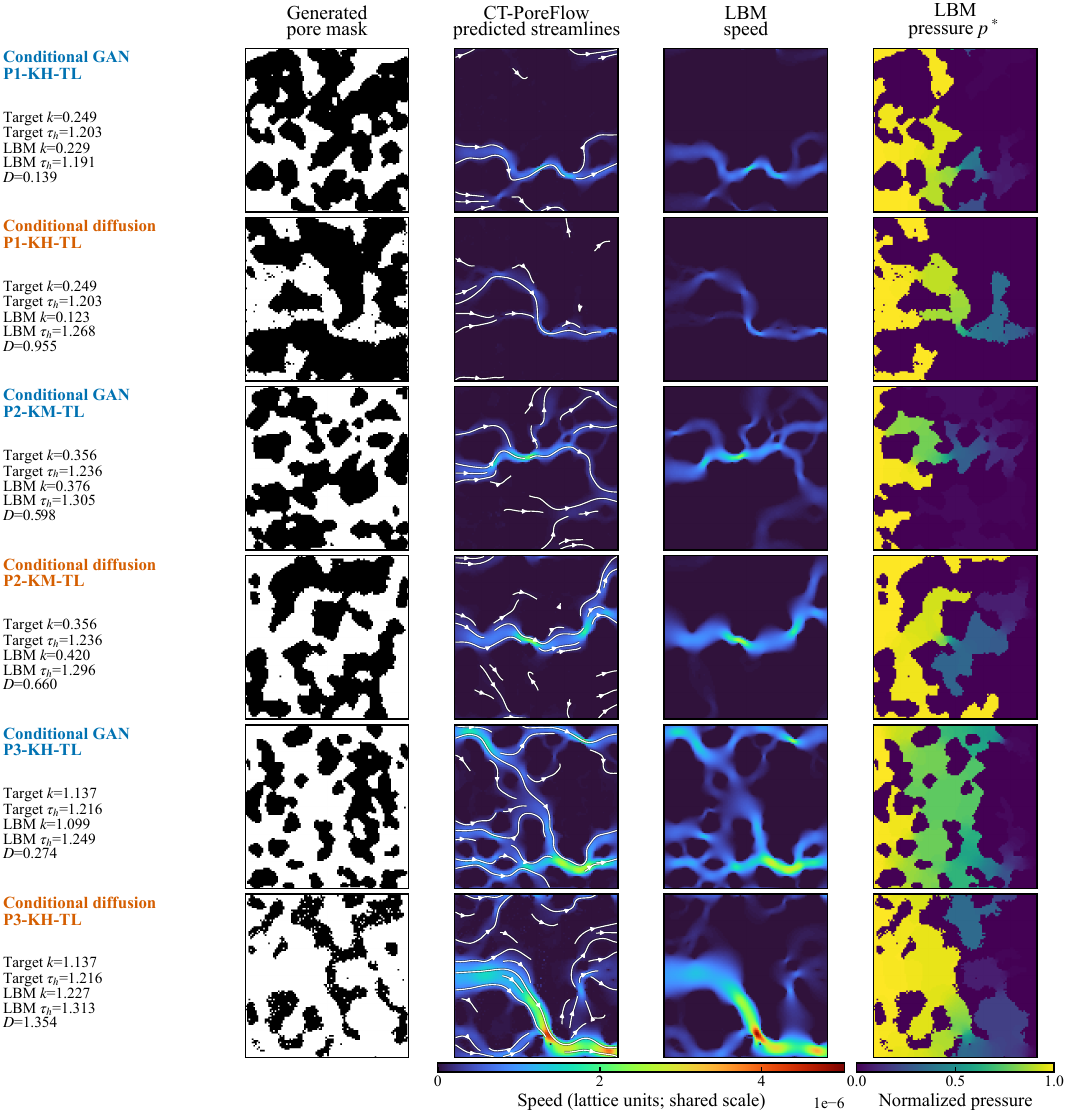}
\caption{Representative QC-passing guided designs at low, middle, and high target porosity.}
\label{fig:generative_mosaic}
\end{figure}

Fig.~\ref{fig:generative_mosaic} presents representative QC-passing guided designs across the low-, middle-, and high-porosity target groups. Together with Table~\ref{tab:generative_lbm} and Fig.~\ref{fig:generative_lbm}, these examples connect the aggregate verification statistics to the corresponding generated geometries, surrogate-predicted flow patterns, and LBM-resolved responses.

\FloatBarrier

\label{sec:design_discussion}
Unlike a scalar property regressor, CT-PoreFlow reconstructs a complete flow
state from which permeability, hydraulic tortuosity, flux balance, and spatially
localised prediction errors can be evaluated simultaneously.

The finite-library benchmark isolates surrogate-ranking error under a fixed level of candidate-set coverage through comparison with the LBM oracle. The fresh-QSGS experiment provides a more stringent solver-based validation by applying the surrogate to previously unseen candidates and subsequently recomputing the selected structures with LBM. Guided screening improved the verified target agreement for all three examined conditions. However, the remaining permeability overshoot for the low-tortuosity target illustrates the coupling between permeability and hydraulic tortuosity, both of which are governed by the organisation of the connected flow pathways. Consequently, matching one transport property does not necessarily yield a simultaneous optimum in the other.

The comparison between the two conditional generators further reveals a trade-off among morphological diversity, hydraulic validity, and target fidelity. Conditional diffusion generated a broader range of binary morphologies but produced fewer through-connected and target-compatible structures, whereas the conditional GAN generated a less diverse population with substantially higher hydraulic validity and closer agreement with the prescribed transport conditions. Morphological novelty should therefore not be interpreted as engineering utility. Including disconnected candidates, failed simulations, random-selection controls,
and computational costs provides a conservative assessment of generative-design
performance.

\section{Limitations and future work} Several limitations define the current scope of this study. First, the benchmark is restricted to two-dimensional QSGS geometries with fixed growth probabilities, lattice resolution, fluid properties, pressure gradient, and flow orientation. Connectivity and convergence filtering also under-represented the lowest-porosity regimes, where 394 generation tasks exhausted the retry limit. Future datasets should therefore include balanced near-percolation samples, alternative morphology-generation rules, variable fluid and forcing parameters, and three-dimensional pore structures. Second, the present validation does not establish universal temporal or morphological generalisation. The main models were trained using one optimisation seed, while the out-of-distribution evaluation was limited to 24 rule-based geometries and eight correlated slices from a single segmented computed tomography volume. Moreover, the directed geodesic channels depend on a prescribed inlet--outlet orientation, and the unseen-time results do not demonstrate temporal extrapolation. Further assessment should include repeated training seeds, independent multi-volume CT datasets, boundary-aware geometry representations, and parameter-conditioned models evaluated under variable hydraulic conditions. Third, the inverse-design results are confined to the training-supported property range and use one seed and one binarisation threshold for each generator without topology repair. Consequently, morphological diversity and surrogate-level target matching do not guarantee hydraulic validity, manufacturability, or engineering performance. Future work should incorporate topology-constrained generation, multi-objective mechanical and transport requirements, three-dimensional LBM verification, and experimental testing of fabricated structures. Until such validation is completed, the generated geometries should be regarded as digitally screened design candidates rather than directly constructible porous-media configurations.

\section{Conclusions}

This study developed a unified and LBM-verified framework that reformulates transient pore-scale flow analysis from a sequence of repeated numerical simulations into a reusable query-time prediction and inverse design problem. The framework integrates the QSGS-Transient-7606 benchmark, the topology- and transport-aware CT-PoreFlow surrogate, engineering-property recovery, property-conditioned candidate generation, and final LBM verification. The main conclusions are as follows.

\begin{enumerate}
\item CT-PoreFlow accurately predicts transient velocity and pressure fields on unseen porous geometries, achieving a velocity relative $L_2$ error of 0.2248 and a terminal permeability error of 12.81\%, improving on Time-U-Net across all paired field and transport metrics under the principal comparison. More importantly, the results show that lower aggregate field error does not necessarily imply better engineering-property recovery. The Direct transient decoder attains the lowest field error before flux calibration, whereas the late-time flux-calibration objective in CT-PoreFlow sacrifices a modest amount of field accuracy to substantially improve terminal flux, permeability, and mass-balance recovery.

\item Total porosity alone is insufficient to characterise pore-scale transport. Connected porosity is strongly associated with permeability, whereas dead-pore volume reduces the hydraulically active pore space. Hydraulic tortuosity is largely independent of connected porosity, indicating that conductance and flow-path complexity represent complementary structural dimensions that should be considered jointly, rather than inferred from porosity alone, when designing or screening porous structures.

\item CT-PoreFlow provides an effective instrument for inverse-design screening, achieving an LBM-verified Success@20 rate of 83.33\% in finite-library retrieval and 72.28\% for guided conditional-GAN designs, both of which exceed their respective random and morphology-only controls. Comparison with the conditional diffusion model shows that greater morphological diversity does not translate into higher connectivity or target fidelity, so diversity and design utility must be reported as separate criteria. 
\end{enumerate}

Overall, the proposed framework provides an efficient and physically accountable link between pore geometry, transient flow evolution, and transport-oriented porous-structure design. These conclusions are established within a fixed two-dimensional QSGS-LBM regime at constant resolution, viscosity, pressure gradient, and boundary orientation, and the image-derived audit reflects a single segmented computed tomography volume rather than an independent multi-sample dataset. Future work should extend the framework to three-dimensional geometries, variable physical conditions, and experimental validation.

\section*{Acknowledgements}
The last author acknowledges the financial support provided by the China Scholarship Council.

\section*{CRediT authorship contribution statement}
\textbf{Yiming Wang:} Methodology, Software, Formal analysis, Data curation, Writing -- original draft, Visualization.
\textbf{Jiale Zhu:} Conceptualization, Methodology, Supervision, Writing -- review \& editing.
\textbf{Zhichen Ye:} Software, Investigation, Validation, Data curation.
\textbf{Yandong Lv:} Investigation, Resources.
\textbf{Shiqi Wang:} Investigation, Formal analysis.
\textbf{Jinlong Liu:} Resources, Validation.
\textbf{Yucheng Fan:} Conceptualization, Methodology, Software, Formal analysis, Visualization, Writing -- review \& editing.

\section*{Declaration of competing interest}
The authors declare that they have no known competing financial
interests or personal relationships that could have appeared to influence
the work reported in this paper.

\section*{Data availability}
The dataset could be found in https://huggingface.co/JonathanWWW/CT-PoreFlow and the code could be found in
https://github.com/wangyiming2/CT-PoreFlow.

\bibliographystyle{cas-model2-names}
\bibliography{refs}

\end{document}